\documentclass[runningheads,a4paper]{llncs}
\usepackage{amsmath}
\usepackage{amsfonts}
\usepackage{amssymb}
\usepackage{graphicx}
\usepackage{enumerate}
\usepackage[capitalise]{cleveref}
\crefname{section}{Sect.}{Sect.}
\crefname{equation}{}{}
\usepackage{bm}
\usepackage{url}
\usepackage{tikz}
\usetikzlibrary{positioning}
\usetikzlibrary{shapes.geometric}
\usepackage{circuitikz}
\usepackage{multirow}
\usepackage{subcaption}
\usepackage{comment}
\usepackage{algorithm}
\usepackage{algpseudocode}
\algnewcommand\algorithmicinput{\textbf{Input:}}
\algnewcommand\Input{\item[\algorithmicinput]}
\algnewcommand\algorithmicoutput{\textbf{Output:}}
\algnewcommand\Output{\item[\algorithmicoutput]}
\algnewcommand\algorithmicforeach{\textbf{for each}}
\algdef{S}[FOR]{ForEach}[1]{\algorithmicforeach\ #1\ \algorithmicdo}

\makeatletter
\AtBeginDocument{%
  \def\doi#1{\url{https://doi.org/#1}}}
\makeatother

\newcommand{\R}[0]{\mathbb{R}}

\newcommand{\ct}[1]{\cos\theta_{#1}}
\newcommand{\st}[1]{\sin\theta_{#1}}

\begin{document}
\title{
  An Effective Trajectory Planning and an Optimized Path Planning for a 6-Degree-of-Freedom Robot Manipulator
}
\titlerunning{An optimized Path Planning for a 6-Degree-of-Freedom Robot Manipulator}
\author{
    Takumu Okazaki\inst{1}\and
    Akira Terui\inst{1}\orcidID{0000-0003-0846-3643} \and
    Masahiko Mikawa\inst{1}\orcidID{0000-0002-2193-3198}
}
\institute{%
University of Tsukuba, Tsukuba, Japan \\
\email{terui@math.tsukuba.ac.jp}\\
\email{mikawa@slis.tsukuba.ac.jp}\\
\url{https://researchmap.jp/aterui}
}

\maketitle

\begin{abstract}
  An effective method for optimizing path planning for a specific model of a 6-degree-of-freedom (6-DOF) robot manipulator is presented
  as part of the motion planning of the manipulator using computer algebra.
  We assume that we are given a path in the form of a set of line segments that the end-effector should follow.
  We also assume that we have a method to solve the inverse kinematic problem of the manipulator at each via-point of the trajectory.
  The proposed method consists of three steps.
  First, we calculate the feasible region of the manipulator under a specific configuration of the end-effector.
  Next, we aim to find a trajectory on the line segments and a sequence of joint configurations the manipulator should follow to move the end-effector along the specified trajectory.
  Finally, we find the optimal combination of solutions to the inverse kinematic problem at each via-point along the trajectory by reducing the problem to a shortest-path problem of the graph and applying Dijkstra's algorithm.
  We show the effectiveness of the proposed method by experiments.
\end{abstract}

\section{Introduction}
\label{sec:introduction}

This paper discusses the motion planning of a 6-Degree-of-Freedom (DOF) robot manipulator.
A manipulator is a robot resembling a human hand, consisting of
\emph{links} that function as a human arm and \emph{joints} as human joints
(see \cref{fig:mycobot-block-diagram}).
Each link is connected by a joint. The first link is connected to the ground, and the last link, called \emph{end-effector}, contains the hand, which can be moved freely.
In this paper, we discuss targeting a manipulator called ``myCobot 280'' \cite{mycobot-280}
(hereafter, we refer to it as ``myCobot'').

In an accompanying paper, we have proposed a method to solve the inverse kinematic problem of myCobot \cite{oka-ter-mik2025a}, which calculates the joint angles of the robot manipulator corresponding to a given end-effector's position and orientation.
In contrast, this paper focuses on the trajectory planning problem and the path planning problems of myCobot.
The trajectory planning problem (see literature, e.g., \cite{gas-bos-lan-vid2012}
and references therein) for robot manipulators is an extension of the inverse
kinematics problem, where the input is expanded from a single position to an entire
trajectory on the given path. Specifically, it involves determining a sequence of
joint configurations that allows the end-effector to move along a given path
between two via-points. In this context, there may exist several solutions to
an inverse kinematic problem at specific via-points along the trajectory.
However, when operating an actual manipulator, it is necessary to select a
single solution from them. The path optimization problem (see literature, e.g.,
\cite{liu-yap-kha2024} and references therein) addresses this by selecting
the optimal solution from the set of possible trajectories obtained from
trajectory planning.

Our previous contributions to trajectory planning are as follows.
For the trajectory planning problem, Yoshizawa et al. \cite{yos-ter-mik2023} proposed a
method to guarantee the existence of the solution of the inverse kinematic problem
at any point on the path
in which the end-effector moves along a straight line between two waypoints,
using Comprehensive Gr\"obner Systems (CGS) \cite{suz-sat2006} and a
quantifier elimination method based on it
(CGS-QE) \cite{fuk-iwa-sat2015}.
Shirato et al. \cite{shi-oka-ter-mik2024} proposed a solution to the trajectory
planning problem in the presence of obstacles located along a straight line between
two waypoints.
They generated obstacle-avoiding trajectories using spline interpolation.
However, with the use of spline interpolation, the generated path might go out
of the feasible region of the manipulator, which may lead to the failure of the
inverse kinematic problem at some via-points along the path.
To address this issue, Hatakeyama et al. \cite{hat-ter-mik2024} proposed a
method for generating
obstacle-avoiding trajectories using B\'ezier curve to guarantee the existence
of the whole path within the feasible region of the end-effector.

In their study, they limited the control to three joints of a 6-DOF robot manipulator,
treating it as a 3-DOF manipulator. As a result, their inverse kinematics formulation
focused solely on the position of the end-effector without considering its orientation.
However, in real-world applications—such as those involving task-oriented robots—it is
often necessary to specify the orientation of the end-effector. Therefore, it is
essential to address the trajectory planning problem for a complete 6-DOF manipulator,
taking into account the end-effector's orientation.

In the context of the path optimization problem, Shirato et al.
\cite{shi-oka-ter-mik2024} applied Dijkstra’s algorithm \cite{dij1959} to the
trajectory planning solutions to find a path from the start to the goal that
minimizes the total joint displacement. However, minimizing the total joint
displacement does not always yield the optimal path. For instance, if a single joint
moves many times, it may lead to excessive wear on that joint, which could shorten
the manipulator’s lifespan. Therefore, it is desirable to flexibly adapt the
optimization criteria according to the specific conditions and requirements of the task.

In this paper, we propose solutions to the trajectory planning and path optimization
problems for the myCobot under the constraint of a fixed end-effector orientation.
We first derive the feasible region of the myCobot under the constraint of a fixed
end-effector orientation. Next, we discuss a solution to the trajectory planning
problem in which the manipulator moves from the start to the goal by dividing the
trajectory into differently spaced segments. One way of dividing the trajectory
consists of the equally spaced segments so that the end-effector moves with continuous speed.
Another division of the trajectory consists of segments such that the end-effector
smoothly starts its motion from a start point, accelerates to a certain speed, and
then decelerates smoothly to stop at the endpoint.
Finally, we propose methods of path optimization using the Dijkstra's algorithm with
several optimization criteria, including minimizing total joint displacement and
comparing them.

This paper is organized as follows.
In \Cref{sec:inverse-kinematics-mycobot}, we briefly review the inverse kinematic computation 
myCobot described in our accompanying paper \cite{oka-ter-mik2025a}.
In \Cref{sec:feasible-region}, we discuss the determination of the feasible region
of the myCobot when the orientation of the end-effector is fixed.
In \Cref{sec:trajectory-planning}, we propose a method for trajectory planning.
In \Cref{sec:path-optimization}, we propose a method for path optimization using
Dijkstra's algorithm.
In \Cref{sec:conclusion}, we discuss findings and future work.

\section{The Inverse Kinematic Problem of myCobot}
\label{sec:inverse-kinematics-mycobot}

\begin{figure}[t]
  \begin{minipage}[b]{0.3\textwidth}
    \centering
    \includegraphics[scale=0.35]{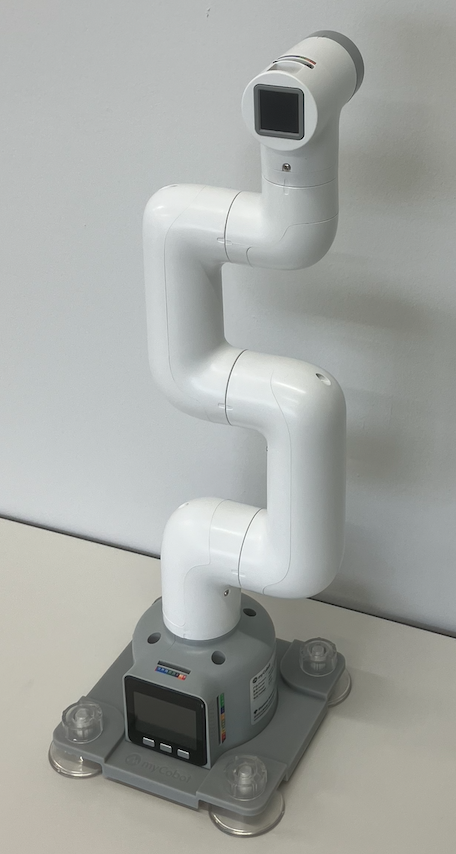}
    \caption{myCobot 280 \cite{mycobot-280}}
    \label{fig:mycobot}
  \end{minipage}
  \hfill
  \begin{minipage}[b]{0.65\textwidth}
  \centering
  \begin{tikzpicture}[scale=0.35]
  \draw (2,1)--(2,5.5);
  \draw (1.5,1.5) rectangle (2.5,3.5); 
  \draw (0.5,6)--(3.5,6)--(3.5,7)--(2,7)--(2,9.5);
  \draw (1,5.5) rectangle (3,6.5); 
  \draw (3.5,10)--(0.5,10)--(0.5,11)--(2,11)--(2,13.5);
  \draw (1,9.5) rectangle (3,10.5); 
  \draw (0.5,14)--(5.5,14);
  \draw (1,13.5) rectangle (3,14.5); 
  \draw (6,12.5)--(6,17.5);
  \draw (5.5,13) rectangle (6.5,15); 
  \draw (6,18) circle(0.5); 
  \draw (6-0.3535,18-0.3535)--(6+0.3535,18+0.3535); 
  \draw (6+0.3535,18-0.3535)--(6-0.3535,18+0.3535); 
  \draw (4,2.5)node{\small Joint 1};
  \draw (4,5)node{\small Joint 2};
  \draw (0.2,9)node{\small Joint 3};
  \draw (4,12.75)node{\small Joint 4};
  \draw (8.5,14)node{\small Joint 5};
  \draw (8.5,17.5)node{\small Joint 6};
  \draw (12,18.5)node{\small The end-effector (Joint 7)};
  \draw (1.5,4.5)node{ $d_1$};
  \draw (1.5,8)node{ $a_2$};
  \draw (1.5,12)node{ $a_3$};
  \draw (4.25,14.5)node{ $d_4$};
  \draw (5.5,16.25)node{ $d_5$};
  \draw (5,18)node{ $d_6$};
  \draw (1.9,1.2) to [out=180,in=90] (1.5,1);
  \draw (1.5,1) to [out=-90,in=180] (2,0.8);
  \draw (2,0.8) to [out=0,in=-90] (2.5,1);
  \draw[arrows = {- stealth}] (2.5,1) to [out=90,in=0] (2.1,1.2);
  \draw (3.1,1)node{ $\theta_1$};

  \draw (0.7,6.1) to [out=90,in=0] (0.5,6.5);
  \draw (0.5,6.5) to [out=180,in=90] (0.3,6);
  \draw (0.3,6) to [out=-90,in=180] (0.5,5.5);
  \draw[arrows = {- stealth}] (0.5,5.5) to [out=0,in=-90] (0.7,5.9);
  \draw (-0.2,6)node{ $\theta_2$};

  \draw (3.3,9.9) to [out=-90,in=180] (3.5,9.5);
  \draw (3.5,9.5) to [out=0,in=-90] (3.7,10);
  \draw (3.7,10) to [out=90,in=0] (3.5,10.5);
  \draw[arrows = {- stealth}] (3.5,10.5) to [out=180,in=90] (3.3,10.1);
  \draw (4.25,10)node{ $\theta_3$};

  \draw (0.7,14.1) to [out=90,in=0] (0.5,14.5);
  \draw (0.5,14.5) to [out=180,in=90] (0.3,14);
  \draw (0.3,14) to [out=-90,in=180] (0.5,13.5);
  \draw[arrows = {- stealth}] (0.5,13.5) to [out=0,in=-90] (0.7,13.9);
  \draw (-0.2,14)node{ $\theta_4$};

  \draw (5.9,12.7) to [out=180,in=90] (5.5,12.5);
  \draw (5.5,12.5) to [out=-90,in=180] (6,12.3);
  \draw (6,12.3) to [out=0,in=-90] (6.5,12.5);
  \draw[arrows = {- stealth}] (6.5,12.5) to [out=90,in=0] (6.1,12.7);
  \draw (6,11.5)node{ $\theta_5$};

  \draw (5.5,18.7) to [out=90,in=180] (6,19.2);
  \draw[arrows = {- stealth}] (6,19.2) to [out=0,in=90] (6.5,18.7);
  \draw (6,20)node{ $\theta_6$};
\end{tikzpicture}
\caption{myCobot with six rotational joints and five links}
\label{fig:mycobot-block-diagram}
\end{minipage}
\end{figure}

In this section, we briefly review the inverse kinematic problem of myCobot
as in our accompanying paper \cite{oka-ter-mik2025a}.

myCobot 280 \cite{mycobot-280} (referred to as ``myCobot'', shown in \Cref{fig:mycobot})
is a 6-DOF robot manipulator with six rotational joints connected in series.
The arm length is 350[mm], and it has a working radius of 280[mm] centered at a height
of approximately 130[mm] from the ground.
myCobot can be controlled using programming languages such as Python, C++, and C\#.
Additionally, it can be controlled using ROS (Robot Operating System) \cite{ROS2},
a standard control environment in robotics.
\Cref{fig:mycobot-block-diagram} shows a block diagram of myCobot.

For the use of the Denavit--Hartenberg convention
\cite{den-har1955} (referred to as ``D-H convention''),
first, we define the symbols necessary for defining the coordinate systems.
Let $\Sigma_i$ be the coordinate system w.r.t.\ Joint $i$, and ${}^ix,{}^iy$,
and ${}^iz$ be
the $x$, $y$, and $z$ axes of $\Sigma_i$, respectively.
Let $\mathcal{O}_i$ be the origin of $\Sigma_i$, and
${}^{i-1}l_i$ be the common normal of the axes ${}^{i-1}z$ and ${}^iz$.

According to the D-H convention, for $i=1,\dots,7$, the coordinate system
$\Sigma_i$ is defined
as a right-handed coordinate system that satisfies the following:
the origin $\mathcal{O}_i$ is placed at Joint $i$;
the axis ${}^iz$ is aligned with the rotation axis of Joint $i$, with the positive
direction pointing towards Joint $i+1$;
the ${}^ix$ and ${}^iy$ axes follow the conventions of the 3D visualization tool
RViz \cite{Rviz} in ROS.
Note that $\Sigma_1$ is treated as the global coordinate system.

Next, the transformation matrix from $\Sigma_{i+1}$ to $\Sigma_{i}$ is defined
with the following parameters.
Note that, for each parameter, the unit of length is expressed in [mm], and the unit of angle is expressed in [rad].
Let $a_i$ be the length of the common normal ${}^{i}l_{i+1}$,
$\alpha_i$ the rotation angle around the ${}^{i+1}x$ axis between the ${}^iz$ and
the ${}^{i+1}z$ axes,
$d_i$ the distance between the common normal ${}^{i}l_{i+1}$ and $\mathcal{O}_i$,
and $\theta_i$ the rotation angle around the ${}^iz$ axis between the common normal
${}^{i}l_{i+1}$ and the ${}^ix$ axis.
Then, we obtain the transformation matrix ${}^iT_{i+1}$ from $\Sigma_{i+1}$ to $\Sigma_i$.

In the coordinate transformation of myCobot in RViz, in addition to the
transformation ${}^iT_{i+1}$,
we add $\delta_i$ as the rotation angle around the
 ${}^{i+1}z$ axis between the ${}^ix$ and the ${}^{i+1}x$ axes.
Thus, we obtain the transformation matrix $A_i$ from $\Sigma_{i+1}$ to $\Sigma_i$
with respect to the coordinate transformation in RViz by adding this effect.
Let $A$ be the transformation matrix
from the end-effector's coordinate system $\Sigma_7$ to the global coordinate system $\Sigma_1$.
Then, $A$ is expressed as
\begin{equation}
  \label{eq:transformation-matrix}
  A=A_1A_2A_3A_4A_5A_6.
\end{equation}


We describe the orientation of the end-effector using Roll-Pitch-Yaw angles \cite{sic-sci-vil-ori2008}.
Let $\bm{l},\bm{m},\bm{n}$ be the unit vectors of the $x,y,z$-axis of the end-effector, respectively, as
\begin{equation}
  \label{eq:orientation}
  \bm{l} = {}^t(l_1,l_2,l_3),\quad
  \bm{m} = {}^t(m_1,m_2,m_3),\quad
  \bm{n} = {}^t(n_1,n_2,n_3).
\end{equation}
Then, the orientation of the end-effector is expressed as
$\begin{pmatrix}
  \bm{l} & \bm{m} & \bm{n}
\end{pmatrix}
$.
Furthermore, let 
\[
  \bm{p}={}^t(p_1,p_2,p_3)
\]
be the position of the end-effector w.r.t. $\Sigma_1$.
Then, for given $\bm{l},\bm{m},\bm{n}$, and $\bm{p}$, the inverse kinematic problem
is to find the joint angles $\theta_1,\dots,\theta_6$.

Based on the above preparations, the inverse kinematic problem of myCobot is solved as follows.
Let $\bm{P}$ be the intersection point of the rotational axes of Joints 4 and 5 as
\begin{equation}
  \label{eq:P}
  \bm{P} = {}^t(x,y,z).
\end{equation}
Note that $\bm{P}$ is at the origin of $\Sigma_5$.
Then, we can obtain $\st{i}$ and $\ct{i}$ ($i=1,\dots,6$) by solving a system of equations $F_1=\cdots=F_{12}=0$, where
\begin{equation}
  \label{eq:angles}
  \begin{split}
  F_1 &= 7318 \st{6}-10^2 (l_1 (-p_1+x)+l_2 (-p_2+y)+l_3 (-p_3+z)),\\
  F_2 &= 7318 \ct{6}-10^2 (m_1 (-p_1+x)+m_2 (-p_2+y)+m_3 (-p_3+z)),\\
  F_3 &= -n_3 \st{5}+\ct{5} (l_3 \ct{6}-m_3 \st{6}),\\
  F_4 &= (\st{5})^2 + (\ct{5})^2-1,\\
  F_5 &= \st{1}+n_1 \st{5}-\ct{5} (l_1 \ct{6}-m_1 \st{6}),\\
  F_6 &= \ct{1}-n_2 \st{5}+\ct{5} (l_2 \ct{6}-m_2 \st{6}),\\
  F_7 &= (\st{3})^2 + (\ct{3})^2-1,\\
  F_8 &= 10^4 x^2+10^4 y^2+(100 z-13156)^2-255799044-211968000 \ct{3},\\
  F_9 &= 13156+11040 \ct{2}+9600 (\ct{2} \ct{3}-\st{2} \st{3})-100 z, \\
  F_{10} &= (\st{2})^2+(\ct{2})^2-1, \\
  F_{11} &= (\st{4})^2+(\ct{4})^2-1, \\
  F_{12} &= (\ct{2} \ct{3}-\st{2} \st{3}) \ct{4}\\
  &\quad -(\st{2} \ct{3}+\ct{2} \st{3}) \st{4}+m_3 \ct{6}+l_3 \st{6}.
  \end{split}
\end{equation}

For deriving the coordinates of $\bm{P}$, we obtain a system of polynomial equations as
\begin{equation}
  \label{eq:CGS}
  \begin{split}
    & n_1(p_1-x)+n_2(p_2-y)+n_3(p_3-z)=d_6,\\
    & (p_1-x)^2+(p_2-y)^2+(p_3-z)^2=d_5^2+d_6^2,\\
    & ((n_1n_2x+(1-n_1^2)y)(p_1-x)\\
    &\quad -(n_1n_2y+(1-n_2^2)x)(p_2-y)-n_3(n_1y-n_2x)(p_3-z) )^2 \\
    &\qquad\qquad =d_4^2(d_5^2n_3^2+(n_2(p_1-x)-n_1(p_2-y))^2),
  \end{split}
\end{equation}
where parameters are $n_1,n_2,n_3$ (direction of the end-effector's $z$-axis),
$p_1,p_2,p_3$ (position of the end-effector).
Note that, in \eqref{eq:CGS}, $d_4,d_5$, and $d_6$ are the lengths of the corresponding links in terms of rational numbers.

To solve \cref{eq:CGS}, one can compute the Gr\"obner basis after substituting the
orientation and position of the end-effector.
However, the computation of the Gr\"obner basis each time after the substitution is
time-consuming. Therefore, by computing the CGS of the ideal generated by the
polynomials in \cref{eq:CGS}, the corresponding Gr\"obner basis can be quickly
obtained once the orientation and position of the end-effector are given,
significantly reducing the time required.

The algorithm for solving the inverse kinematics problem, summarized above,
is as follows
\cite[Algorithm 1]{oka-ter-mik2025a}. First, the CGS for the system of polynomial
equations \Cref{eq:CGS} ---whose coefficients have the position and orientation
of the end-effector as parameters---is precomputed.
Given the input position and orientation of the end-effector, the corresponding
segment in the CGS for \Cref{eq:CGS} is identified.
Then, using the Gr\"obner basis associated with the selected CGS segment, the intersection point
$\bm{P}$ in \Cref{eq:P} is computed. Subsequently, the coordinates of the
intersection point
$\bm{P}$ are substituted into the Gr\"obner basis of the polynomial system
\Cref{eq:angles} to determine the joint angles.

\section{Estimating the Feasible Region}
\label{sec:feasible-region}

According to the experiments we conducted in the accompanying paper
\cite{oka-ter-mik2025a}, the results suggest that the existence of solutions to
the inverse kinematics problem of myCobot depends on both the position and
orientation of the end-effector.
Thus, we call the ``feasible region'' of the end-effector the one for a given
(fixed) orientation of the end-effector.
We first examine the feasible region of the end-effector because
the whole path of the end-effector should be contained within the feasible region.

Furthermore, we assume that the orientation of the end-effector is constrained
to a predetermined configuration.
In the following, we determine the feasible region by considering the cases
where $n_3=\pm 1$, $n_3=0$, and $n_3\neq 0, \pm1$ separately.

\subsection{In the Case $n_3=\pm 1$}
\label{sec:pm1}

If we assume $n_3=\pm 1$, we have $n_1=n_2=0$.
Then, \eqref{eq:CGS} becomes
\begin{equation}
  \label{eq:CGS_n3_1}
  \begin{split}
  (p_3-z)^2=d_6^2,\quad
  (p_1-x)^2+(p_2-y)^2 = d_5^2,\quad
  (yp_1-xp_2)^2 = d_4^2d_5^2.
  \end{split}
\end{equation}
By the first equation in \eqref{eq:CGS_n3_1}, we always have a real number solution
in $z$ for any $p_3 \in \R$. Thus, $p_3$ is any real number.
By the third equation, we have $yp_1-xp_2 = \pm d_4d_5$, which cannot be satisfied
for $p_1=p_2=0$.
Thus, we assume that $p_1\ne 0$ and solve it for $y$ as $y=\frac{\pm d_4d_5+xp_2}{p_1}$,
and by putting it into the second equation of \eqref{eq:CGS_n3_1}, we have
\begin{equation}
  \label{eq:real-root-1}
  \left( 1+\frac{p_2^2}{p_1^2} \right) x^2
  -2\left(\frac{p_2^2}{p_1}\mp\frac{d_4d_5p_2}{p_1^2}+p_1 \right) x
  -d_5^2+\frac{d_4^2d_5^2}{p_1^2}+p_1^2\mp\frac{2d_4d_5p_2}{p_1}+p_2^2 = 0.
\end{equation}
By calculating the resultant of \eqref{eq:real-root-1}, we see that we have real
number solutions in $x$ and $y$ if
$\frac{d_5^2(-d_4^2+p_1^2+p_2^2)}{p_1^2} \ge 0$,
that is $p_1^2+p_2^2 \ge d_4^2$ and $p_1\ne 0$.
Similarly, by assuming that $p_2\ne 0$ and solving it for $y$ as $y=\frac{\pm
d_4d_5+yp_1}{p_2}$,
and by putting it into the second equation of \eqref{eq:CGS_n3_1}, we have
a formula by interchanging $p_1$ and $p_2$ in \eqref{eq:real-root-1}.
By calculating its resultant, we see that we have real number solutions in $x$
and $y$ if $p_1^2+p_2^2 \ge d_4^2$ and $p_2\ne 0$.
Summarizing the above, we have the feasible region as
\begin{equation}
  \label{eq:feasible-region-1}
  p_1^2+p_2^2\ge d_4^2, \quad p_3\in\R.
\end{equation}

\subsection{In the Case $n_3=0$}
\label{sec:n3=0}

If we assume $n_3=0$, we have the following system of polynomial equations
on the position of $\bm{P}$ (see \cite{oka-ter-mik2025a}):
\begin{equation}
  \label{eq:CGS_n3_0_1}
  \begin{split}
  n_1(p_1-x)+n_2(p_2-y)&=d_6,\\
  (p_1-x)^2+(p_2-y)^2+(p_3-z)^2&=d_5^2+d_6^2,\\
  n_2(p_1-x)-n_1(p_2-y) &= 0.
  \end{split}
\end{equation}
From the first and the third equations in \eqref{eq:CGS_n3_0_1}, we have
\begin{equation}
  \label{eq:n3-xy}
  \begin{split}
  (n_1^2+n_2^2)(p_1-x) &= p_1-x = n_1d_6,\quad
  (n_1^2+n_2^2)(p_2-y) = p_2-y = n_2d_6,
  \end{split}
\end{equation}
which shows that the set of values of $p_1$ and $p_2$ that yield real solutions
for $x$ and $y$ is $\R^2$.
Putting \eqref{eq:n3-xy} into the second equation in \eqref{eq:CGS_n3_0_1}, the
set of values of $p_3$ that yields real solutions for $z$ is $\R$.
Thus, in the case $n_3=0$, the feasible region is $\R^3$.

\subsection{In the Other Cases}
\label{sec:other-case}

To determine the feasible region for the case $n_3\ne 0, \pm 1$, it suffices to
derive the constraints on $p_1,p_2$, and $p_3$ such that the parameterized system
of equations, obtained by substituting the desired orientation $(n_1,n_2,n_3)$
into \eqref{eq:CGS}, admits real solutions.
We have attempted to derive the constraints using quantifier elimination in the
computer algebra system Mathematica 14.0; however, the computation was not complete
even after running for at least two days, and the desired result could not be obtained.

\section{Trajectory Planning}
\label{sec:trajectory-planning}

In this paper, we derive a solution to the trajectory planning problem for the myCobot,
where the end-effector is constrained to a specific orientation in each segment and
moves along a linear path from the initial position to the target position. The
trajectory is first divided into $T$
segments, and $T+1$ via-points---including the initial and the target positions
---are computed. The trajectory planning problem is then solved by solving the
inverse kinematics for the coordinates of each via-point.

First, the coordinates of the end-effector, the initial position, and the target
position are defined as
$(x,y,z)$, $(x_0,y_0,z_0)$, and $(x_T,y_T,z_T)$, respectively.
By using a parameter $s\in[0,1]$, the coordinates of the end-effector are represented
as a function of $s$ as
\begin{equation}
  \label{eq:coordinates}
  f(s) = {}^t(x,y,z)
  = {}^t(x_0(1-s)+x_Ts, y_0(1-s)+y_Ts, z_0(1-s)+z_Ts).
\end{equation}

In the following, we describe two methods for trajectory generation: 1) uniform motion along the path as a function of time and 2) smooth motion along the path as a function of time.

\subsection{Generating a Trajectory for Uniform Motion Along the Path as a Function of Time}
\label{sec:trajectory-uniformly}

We generate a trajectory that moves along the path at a constant speed using
a time-dependent function as follows.
For time $t=0,1,2,\dots,T$, the parameter $s$ is represented as $s=t/T$.
In this case, the end-effector is at the initial position for $t=0$,
at the final position for $t=T$,
and at one of the division points for all other values of $t$.

\subsection{Generating a Trajectory for Smooth Motion Along the Path as a Function of Time}
\label{sec:trajectory-smooth}

We generate a trajectory that moves smoothly as a function of time, as follows.
When the end-effector begins to move from rest, both its velocity and acceleration
start from zero and increase smoothly.
Furthermore, prior to coming to a complete stop, the end-effector's velocity
and acceleration both decrease continuously, with the acceleration reaching zero
at the moment of stopping.
For expressing the coordinates of the end-effector,
let us represent the parameter $s$ as $s(t)$, which is a function of time $t$,
and let $\dot{s}$ and $\ddot{s}$ be the first and the second derivative of $s$,
respectively.
Then we have
\begin{equation}
  \label{eq:derivatives}
  \begin{split}
    \frac{df}{dt} &={}^t((x_T-x_0)\dot{s},(y_T-y_0)\dot{s},(z_T-z_0)\dot{s}), \\
    \frac{d^2f}{dt^2} &= {}^t((x_T-x_0)\ddot{s},(y_T-y_0)\ddot{s}, (z_T-z_0)\ddot{s}),
  \end{split}
\end{equation}
thus the velocity and the acceleration are proportional to $\dot{s}$ and $\ddot{s}$,
respectively.

In the case the end-effector moves at a constant speed as shown in
\cref{sec:trajectory-uniformly}, the acceleration changes suddenly at the beginning
and end of the motion.
However, it is desirable for the manipulator's motion to be smooth.
Therefore, for each time step $t=0,\dots,T$, the function $s(t)$ is required to satisfy
\begin{equation}
  \label{eq:s-constraints}
  s(0)=0,\quad s(T)=1,\quad \dot{s}(0)=\dot{s}(T)=\ddot{s}(0)=\ddot{s}(T)=0.
\end{equation}

A function $s(t)$ satisfying \eqref{eq:s-constraints} can be expressed as a polynomial
of degree 5 in $t$ \cite{lyn-par2017} as follows.
Let
\begin{equation}
  \label{eq:s}
  s(t) = a_5\left(\frac{t}{T}\right)^5+a_4\left(\frac{t}{T}\right)^4+a_3\left(\frac{t}{T}\right)^3+a_2\left(\frac{t}{T}\right)^2+a_1\left(\frac{t}{T}\right)+a_0,
\end{equation}
where $a_i$ ($i=0,\dots,5$) are real numbers.
Then, $\dot{s}(t)$ and $\ddot{s}(t)$ become
\begin{equation}
  \label{eq:dotddot}
  \begin{split}
    \dot{s}(t) &= \frac{5a_5}{T^5}t^4+\frac{4a_4}{T^4}t^3+\frac{3a_3}{T^3}t^2+\frac{2a_2}{T^2}t+\frac{a_1}{T},\\
    \ddot{s}(t) &= \frac{20a_5}{T^5}t^3+\frac{12a_4}{T^4}t^2+\frac{6a_3}{T^3}t+\frac{2a_2}{T^2}.
  \end{split}
\end{equation}
By the constraints \eqref{eq:s-constraints}, we see that $a_0=a_1=a_2=0$ and $a_3,a_4,a_5$ satisfy the following system of linear equations:
\begin{equation}
  \label{eq:a-system}
  a_3+a_4+a_5-1=0, \quad
  3a_3+4a_4+5a_5=0, \quad
  2(3a_3+6a_4+10a_5)=0.
\end{equation}
By solving \eqref{eq:a-system}, we have $a_5=6$, $a_4=-15$, $a_3=10$, which derives that
\[
s(t) =\frac{6}{T^5}t^5-\frac{15}{T^4}t^4+\frac{10}{T^3}t^3.
\]

\subsection{Algorithm}
\label{sec:trajectory-algorithm}

The algorithm for solving the trajectory planning problem is shown in
\Cref{alg:trajectory}.
Note that the algorithm can solve the problem with different configurations of
the end-effector  given for each segment of the path.
Furthermore, the algorithm takes as input a sequence of points referred to as
``waypoints'' to define the trajectory. The trajectory is constructed as a set
of line segments connecting each pair of adjacent waypoints.

\begin{algorithm}[t]
  \caption{Solving the trajectory planning problem for myCobot}
  \label{alg:trajectory}
  \begin{algorithmic}[1]
    \Input{The set of waypoints along the path $\{(x_i,y_i,z_i)\mid i=0,\dots,N-1\}$;
     The set of orientations of the end-effector at each segment of the path
     $\{(\alpha_i,\beta_i,\gamma_i)\mid i=1,\dots,N-1\}$;
     The number of subdivisions $T$ for each segment of the path;}
    \Output{A set $J$ of tuples of joint angles $(\theta_1,\dots,\theta_6)$;}
    \State{$O\gets\emptyset$; $K\gets\emptyset$; $J\gets\emptyset$;}
    \For{$i\in[0,\dots,N-1]$}
      \For{$t \in [0,\dots,T]$}
        \If{(the end-effector moves along the path at a constant speed)}
          \State{$s\gets t/T$;}
        \Else
          \State{$s\gets\frac{6}{T^5}t^5-\frac{15}{T^4}t^4+\frac{10}{T^3}t^3$;}
        \EndIf
        \State{$(x_t,y_t,z_t)\gets(x_j+(x_{j+1}-x_j)s , y_j+(y_{j+1}-y_j)s, z_j+(z_{j+1}-z_j)s)$};
        \State{$K\gets K\cup\{(x_t,y_t,z_t)\}$; $O\gets O\cup\{(\alpha_i,\beta_i,\gamma_i)\}$;}
      \EndFor
    \EndFor
    \While{$K\ne\emptyset$}
      \State{Extract the first element $(x,y,z)$ from $K$ and the first element
       $(\alpha,\beta,\gamma)$ from $O$;}
      \State{Solve the inverse kinematics problem using the end-effector's position $(x,y,z)$ and orientation $(\alpha,\beta,\gamma)$;
      Let $(\theta_1,\theta_2,\theta_3,\theta_4,\theta_5,\theta_6)$ denote the joint angles obtained as the output;}
      \Comment{\cite[Algorithm 1]{oka-ter-mik2025a}}
      \State{$J\gets J\cup\{(\theta_1,\theta_2,\theta_3,\theta_4,\theta_5,\theta_6)\}$;}
    \EndWhile
    \State{\Return $J$;}
  \end{algorithmic}
\end{algorithm}

\begin{remark}
  \Cref{alg:trajectory} works properly, as follows. First, for each given line segment,
  it
  calculates a series of via-points $K$ on which the end-effector should follow, and a
  series of
  orientations of the end-effector $O$ for each line segment.
  Then, it solves the inverse kinematic problem for each position in $K$ and the
  orientation in
  $O$, and generates a series of configurations of the joints.
\end{remark}

\subsection{Experiments}
\label{sec:trajectory-experiments}

We show the result of experiments using \Cref{alg:trajectory}.
The computing environment for the experiments in this paper is as follows:
Intel Xeon Silver 4210 3.2 GHz,
RAM 256 GB,
Linux Kernel 5.4.0,
Risa/Asir Version 20230315\footnote{The functionality for computing numerical roots
of univariate polynomial equations utilizes the capabilities of the computer
algebra system PARI/GP 2.13.1
\cite{PARI2.13.1}, which is invoked as a built-in function from Risa/Asir.}.

First, in \Cref{alg:trajectory}, trajectory planning is performed under the assumption
that the end-effector maintains a fixed orientation identical to that of the global
coordinate system throughout all segments.
Specifically, the orientation is defined by $\bm{l}={}^t(1,0,0)$,
$\bm{m}={}^t(0,1,0)$, and $\bm{n}={}^t(0,0,1)$.
Note that, in this case, the feasible region of the end-effector becomes
\eqref{eq:feasible-region-1}. By putting D-H parameters, we obtain the constraints
for the feasible region as
$p_1^2+p_2^2\ge 64.62^2$, $p_3\in\R$.

\begin{figure}[t]
\centering
\begin{minipage}[b]{0.47\columnwidth}
    \centering
    \includegraphics[width=1\columnwidth]{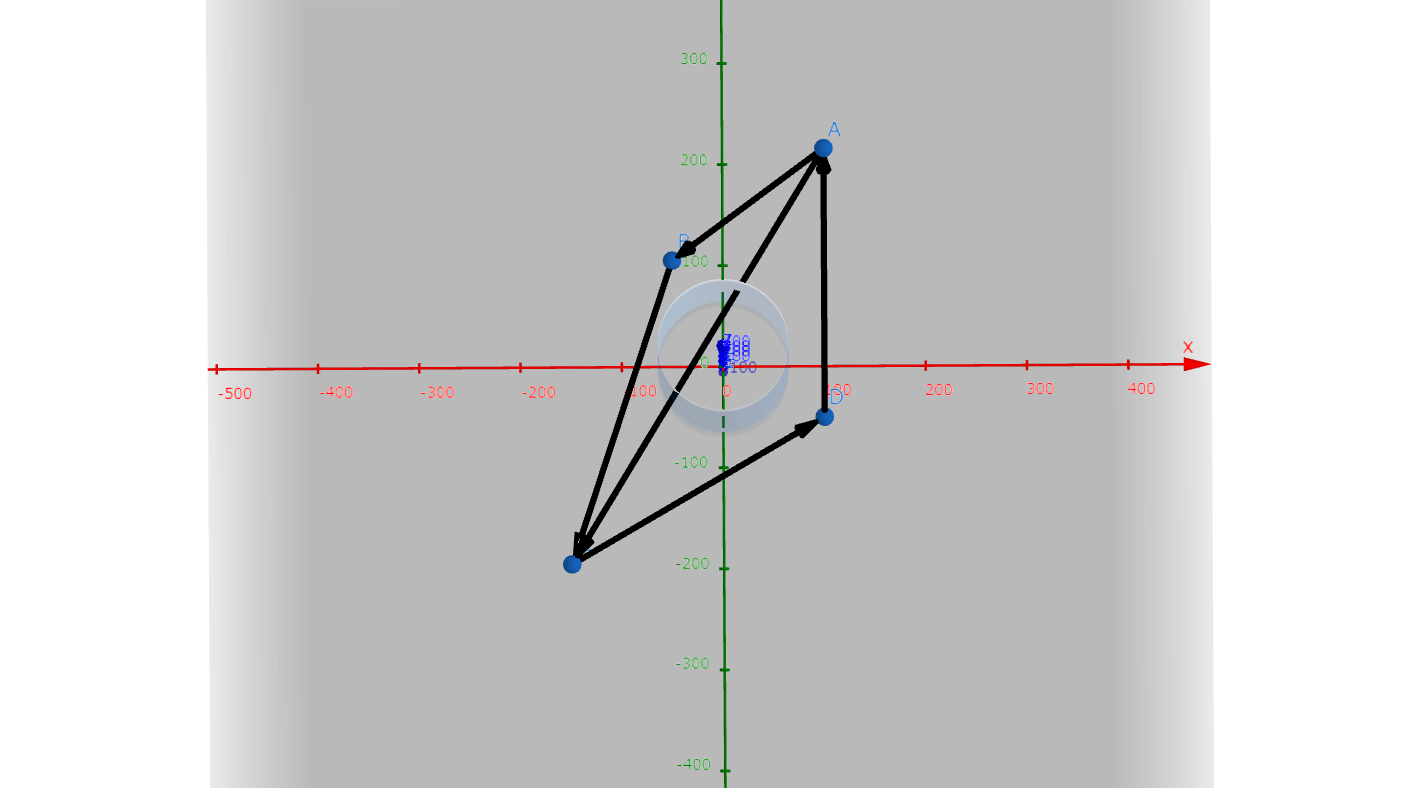}
    \caption{The trajectories and feasible regions for Test 1 through 5 (A view from the positive direction of the $z$-axis)}
    \label{fig:tra-z}
\end{minipage}
\hfill
\begin{minipage}[b]{0.47\columnwidth}
    \centering
    \includegraphics[width=1\columnwidth]{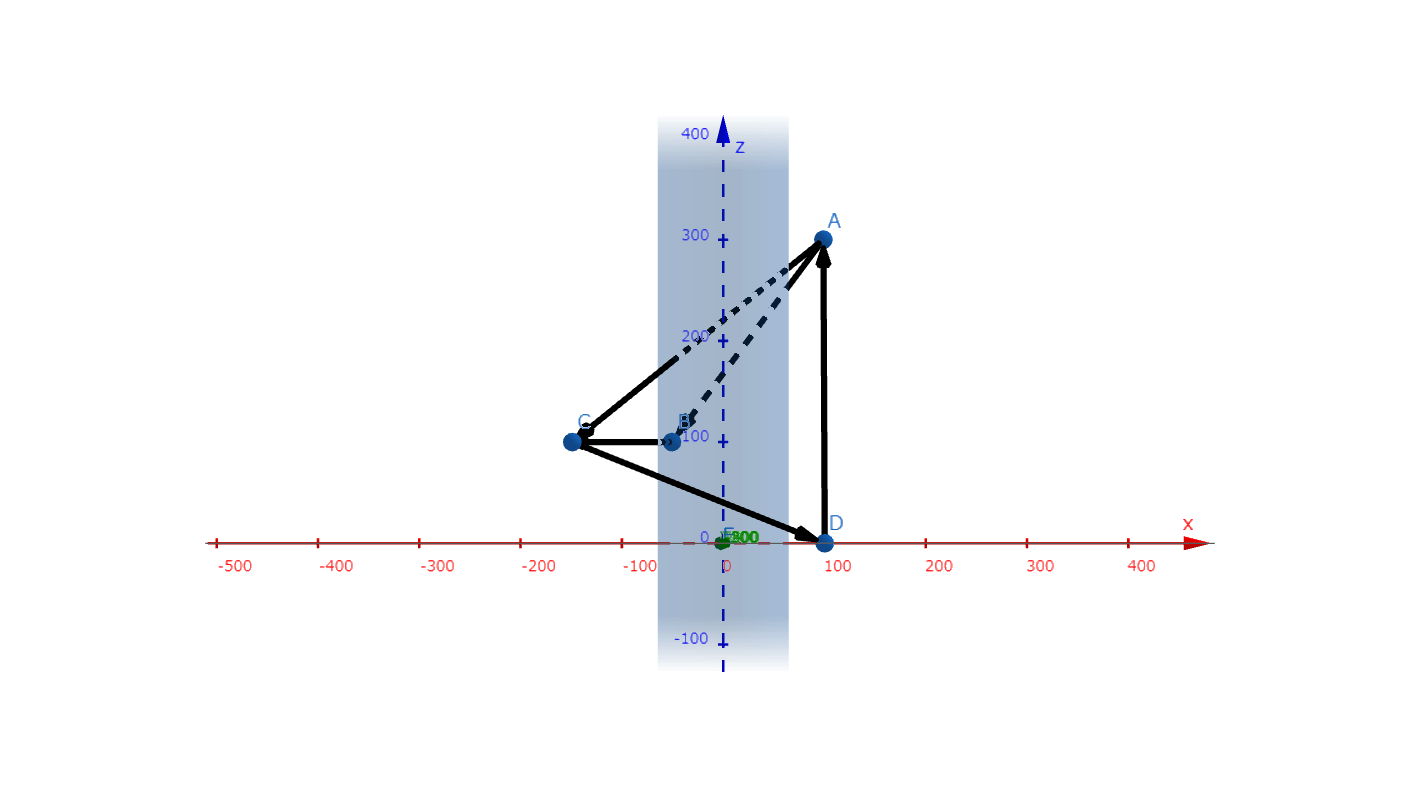}
    \caption{The trajectories and feasible regions for Test 1 through 5 (A view from the positive direction of the $y$-axis)}
    \label{fig:tra-y}
\end{minipage}
\end{figure}

In the experiments, the following waypoints are given:
$A(100, 200, 300)$,
\linebreak
$B(-50, 100, 100)$,
$C(-150, -200, 100)$, $D(100,-50,0)$,
and the following line segments are given as the path:
Test 1: $AB$, Test 2: $BC$,
Test 3: $CD$, Test 4: $DA$,
Test 5: $AC$, Test 6: quadrilateral $ABCD$.
\cref{fig:tra-z,fig:tra-y} show the trajectories and the feasible region.
The cylinder in the figures represents the boundary of the feasible region:
$p_1^2+p_2^2=64.62^2$.
\cref{fig:tra-z} shows that the trajectories of
Test~1 through 4 lies entirely within the feasible region, whereas a portion
of the trajectory in Test~5 extends outside the
\begin{figure}[t]
  \begin{minipage}[t]{0.45\hsize}
    \centering
    \includegraphics[keepaspectratio, scale=0.10]{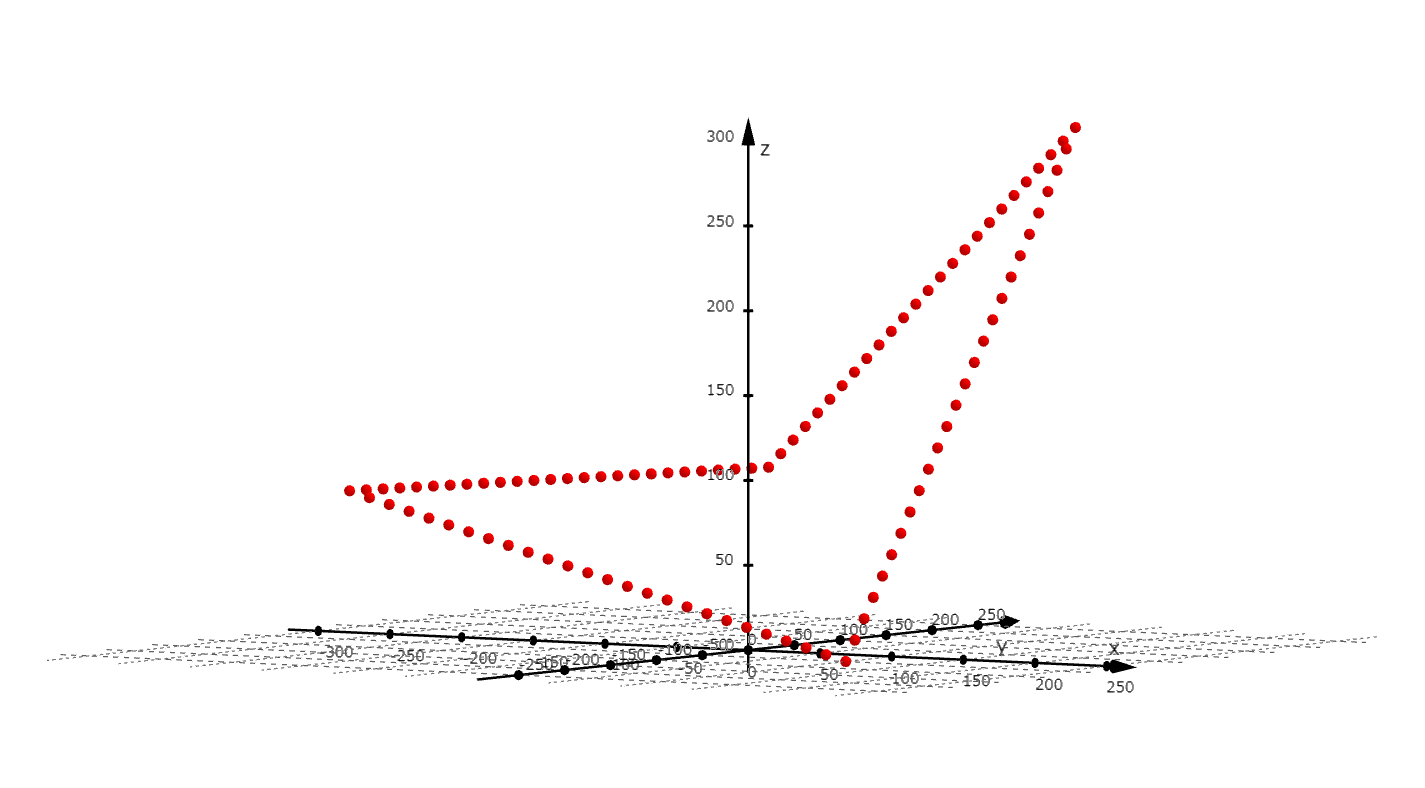}
    \caption{The trajectory generated for uniform motion with $T=25$}
    \label{fig:eq-25}
  \end{minipage}
  \hfill
  \begin{minipage}[t]{0.45\hsize}
    \centering
    \includegraphics[keepaspectratio, scale=0.10]{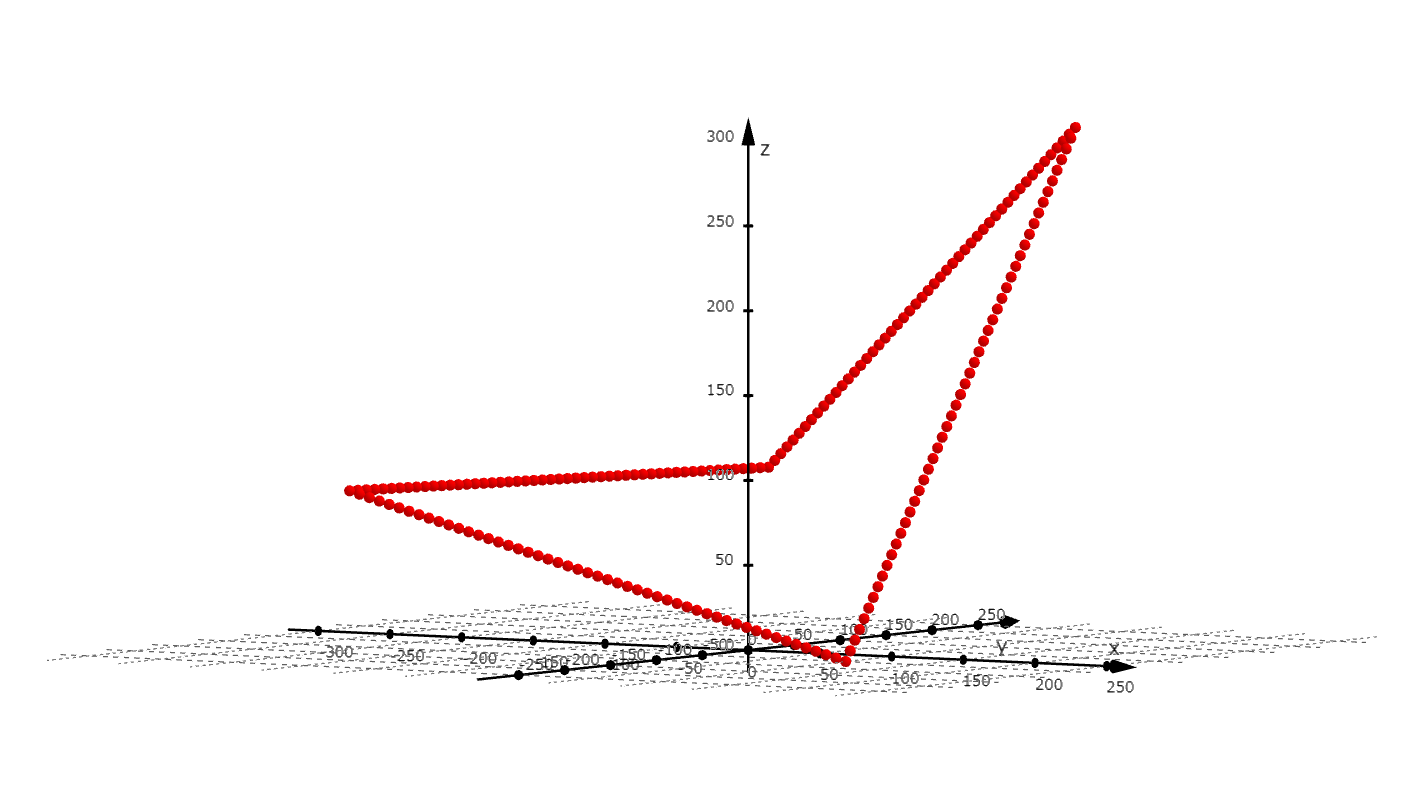}
    \caption{The trajectory generated for uniform motion with $T=50$}
    \label{fig:eq-50}
  \end{minipage} \\
\end{figure}
\begin{figure}[H]
\centering
  \begin{minipage}[t]{0.45\hsize}
    \centering
    \includegraphics[keepaspectratio, scale=0.10]{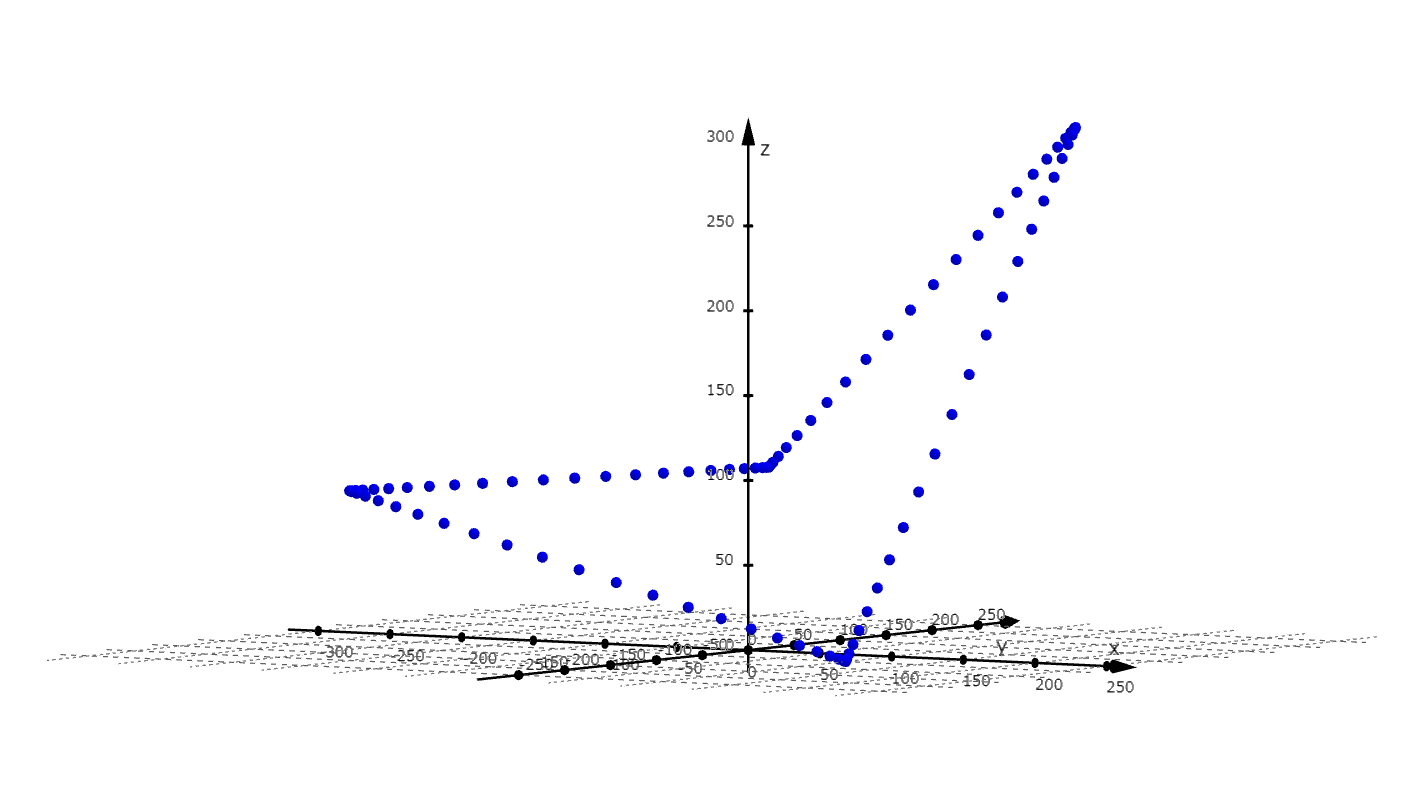}
    \caption{The trajectory generated for smooth motion with $T=25$}
    \label{fig:sm-25}
  \end{minipage}
  \hfill
  \begin{minipage}[t]{0.45\hsize}
    \centering
    \includegraphics[keepaspectratio, scale=0.10]{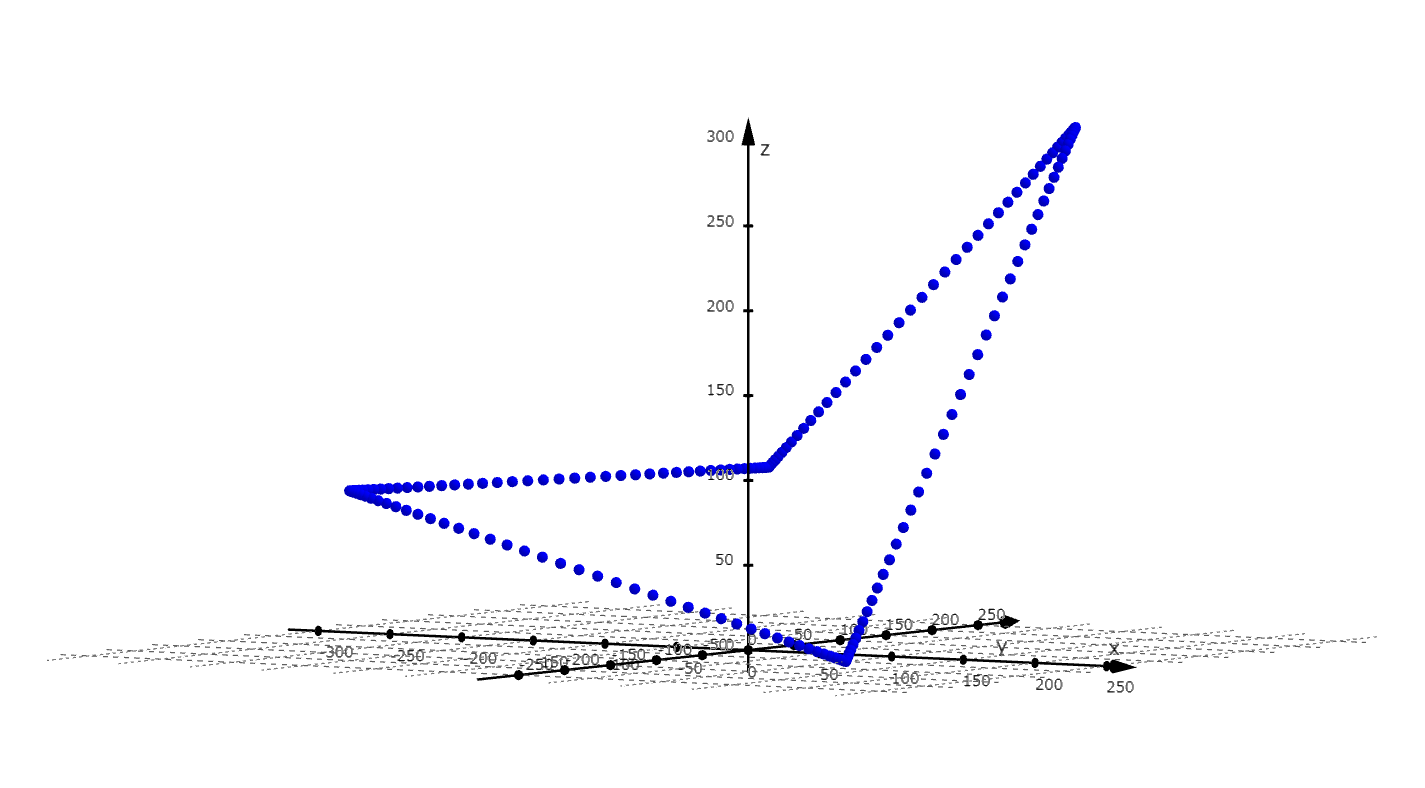}
    \caption{The trajectory generated for smooth motion with $T=50$}
    \label{fig:sm-50}
  \end{minipage}
\end{figure}
\noindent feasible region.
Note that Test~6 represents a single trajectory that combines those of Tests~1
through 4.

For Tests~1 through~6, the computing times for a total of 24 trajectory planning cases
were measured for generating trajectories for both uniform motions (see \cref{sec:trajectory-uniformly}) and smooth motion (see \cref{sec:trajectory-smooth}),
with $T=25$ and $T=50$ for each case.
The via-points generated in Test~6 are illustrated in
\cref{fig:eq-25,fig:eq-50,fig:sm-25,fig:sm-50}.
Tests~1 through~4 each correspond to one side of the trajectory shown in these
figures.

\cref{tab:trajectory} shows the results of the experiments.
Each computation time (in seconds) represents the average over 10 executions
of trajectory generation for the corresponding path.
The columns with ``Uniform motion'' are the results for generating trajectories
for uniform motion (\cref{sec:trajectory-uniformly}), while the columns with
``Smooth motion'' is the result of
generating trajectories for smooth motion (\cref{sec:trajectory-smooth}).
Based on the experimental results, there appears to be no significant difference
in computing time between the proposed method for uniform motion and that for
smooth motion.
As noted above, a portion of the trajectory in Test~5 lies outside the feasible
region, and thus, it was not possible to compute all trajectories for this case.
Accordingly, the computation time for Test~5 is shorter compared to that of the
other tests.

We see that, since Test~6 is a combined trajectory that integrates Tests~1 through~4,
its computing time is approximately equal to the sum of the computing times of Tests~1
to~4.
It is also observed that the computing time for $T=50$ is roughly twice that for $T=25$.
Furthermore, the number of via-points in Test~6 is approximately four times that
of Tests~1 through~4 combined, and the computing time increases by a similar factor.
These observations suggest that the computation time is approximately proportional
to $T$.
\begin{table}[t]
  \centering
  \begin{tabular}{c|cc|cc}
  \hline
  \multirow{2}{*}{Test} & \multicolumn{2}{|c|}{Uniform motion} &
  \multicolumn{2}{|c}{Smooth motion} \\ \cline{2-5}
  & $T=25$  & $T=50$ & $T=25$ & $T=50$ \\
  \hline
  1 & 2.551 & 4.515 & 2.312 & 4.489 \\ 
  2 & 2.447 & 4.785 & 2.425 & 4.753 \\ 
  3 & 2.297 & 4.492 & 2.306 & 4.516 \\ 
  4 & 2.338 & 4.631 & 2.290 & 4.495 \\ 
  5 & 2.023 & 4.016 & 2.112 & 4.076 \\ 
  6 & 9.413 & 18.700 & 9.384 & 18.541 \\ \hline
  \end{tabular}
  \caption{Average computing time [s] of Tests~1 through 6}
  \label{tab:trajectory}
\end{table}

\begin{figure}[t]
  \centering
  \begin{minipage}[b]{0.45\columnwidth}
      \centering
      \includegraphics[scale=0.4]{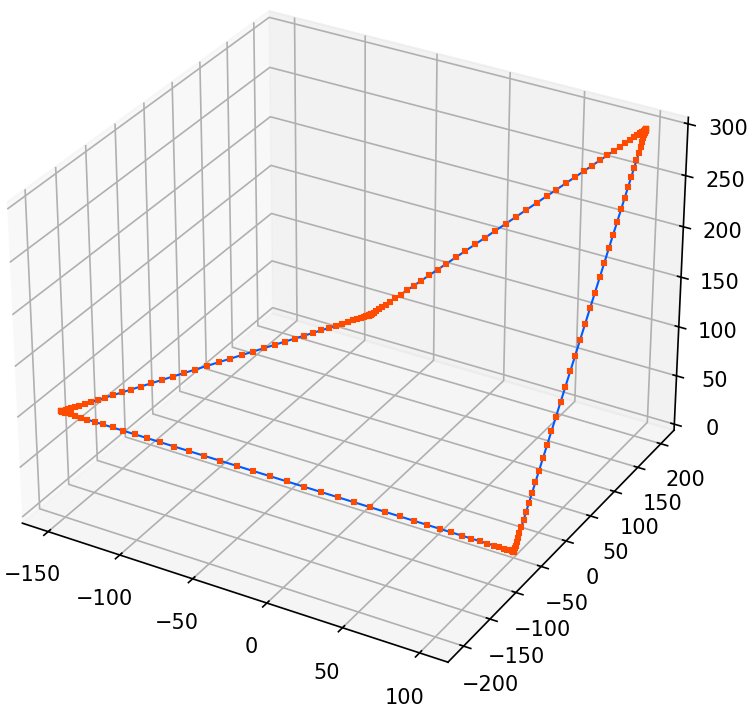}
  \end{minipage}
  \hfill
  \begin{minipage}[b]{0.45\columnwidth}
      \centering
      \includegraphics[scale=0.4]{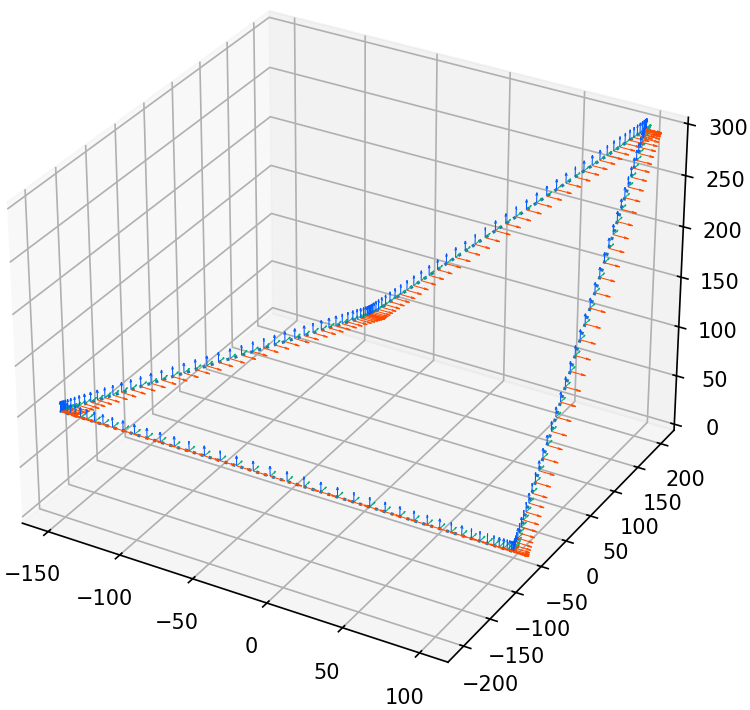}
  \end{minipage}
  \caption{The position and orientation of the end-effector for Test 6 with trajectories for smooth motion and $T=50$}
  \label{fig:Sm-50-all}
\end{figure}

\cref{fig:Sm-50-all} shows the plotted trajectory of the actual position and orientation
of the end-effector during the execution of Test~6 with a trajectory for smooth motion
(\cref{sec:trajectory-smooth}) and $T=50$.
The left figure shows the positions of the end-effector derived from the solutions
to the inverse kinematics problem at each waypoint.
In contrast, the right figure shows the corresponding orientations of the end-effector.
Note that in cases where multiple solutions to the inverse kinematics problem exist
at a given point, one solution is randomly selected from among them.
From the figure, we see that the end-effector closely follows the desired trajectory,
and its orientation is maintained consistently.

In the next experiment,
for each segment of the path, the orientation of the end-effector is modified,
and the corresponding trajectory planning problem is formulated and solved.
The following waypoints are given:
$A'(100, 100, 200)$, $B'(-100, 100, 100)$, $C'(-100, -100, 0)$, $D'(100,-100,0)$,
$E'(100,100,100)$,
and the following line segments are given as the path:
Test 7: pentagon $A'B'C'D'E'$.
Furthermore, the orientation of the end-effector is given for each line segment
as follows:
$A'B'$: $(\pi/2,\pi/2,\pi/2)$,
$B'C'$: $(\-\pi/2,0,-\pi/2)$,
$C'D'$: $(-\pi/2,-\pi/2,-\pi/2)$,
$D'E'$ $(\pi/2,0,\pi/2)$,
$E'A'$ $(0,0,0)$.
Note that the orientation of the end-effector satisfies $n_3=0$ in $E'A'$ and
$n_3\ne 0$ in the other line segments.

\begin{figure}[t]
  \centering
  \begin{minipage}[b]{0.45\columnwidth}
      \centering
      \includegraphics[scale=0.4]{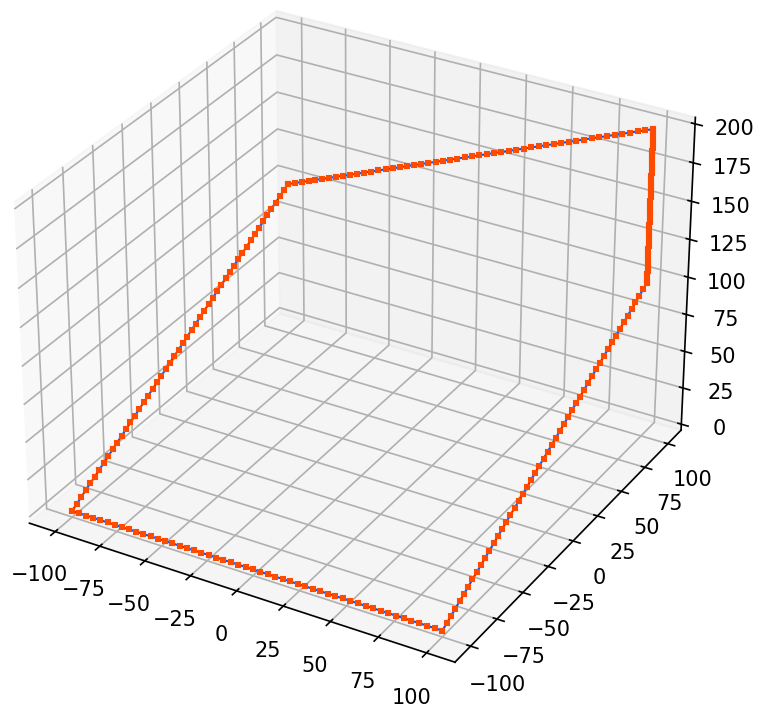}
  \end{minipage}
  \hfill
  \begin{minipage}[b]{0.45\columnwidth}
      \centering
      \includegraphics[scale=0.4]{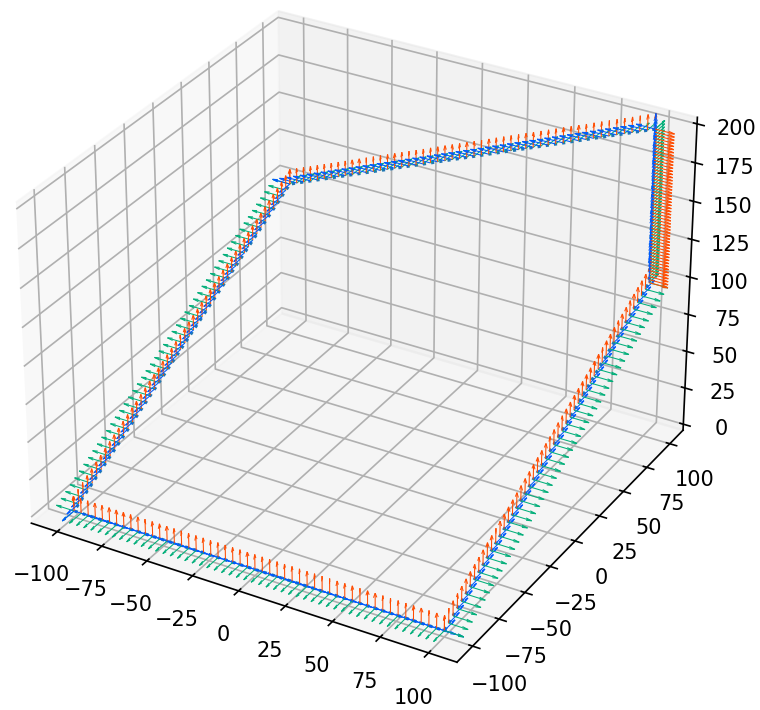}
  \end{minipage}
  \caption{The position and orientation of the end-effector for Test 7 with trajectories for uniform motion and $T=50$}
  \label{fig:Eq-50-ori}
\end{figure}

\cref{fig:Eq-50-ori} shows the trajectories of the position and the orientation
of the end-effector for Test 7, with trajectories generated for uniform motion
(\cref{sec:trajectory-uniformly}) and $T=50$.
Note that the configurations of the figures on the right and left are the same
as those in
\cref{fig:Sm-50-all}.
The computing time was approximately 28.709 seconds.
This experiment also shows that the position and orientation of the end-effector
at each via-point are very close to the given values.

Furthermore, when attempting to solve Tests 1 through 6 using a fixed end-effector
orientation such that $n_3=0$, certain orientations of the end-effector were found
for which no solution to the inverse kinematics problem exists, although the analysis
in \cref{sec:n3=0} shows that the feasible region of the end-effector is $\R^3$
for $n_3=0$.
This suggests that the feasible region obtained in \cref{sec:feasible-region}
contains the actual feasible region, but may not necessarily coincide with it.

\section{Path Optimization}
\label{sec:path-optimization}

The trajectory obtained in \cref{sec:trajectory-planning} may have multiple inverse kinematics solutions at each via-point. Therefore, to execute the trajectory of the
myCobot, it is necessary to select one inverse kinematics solution at each via-point to generate a feasible path.

\begin{figure}[t]
  \centering
  \includegraphics[scale=0.2]{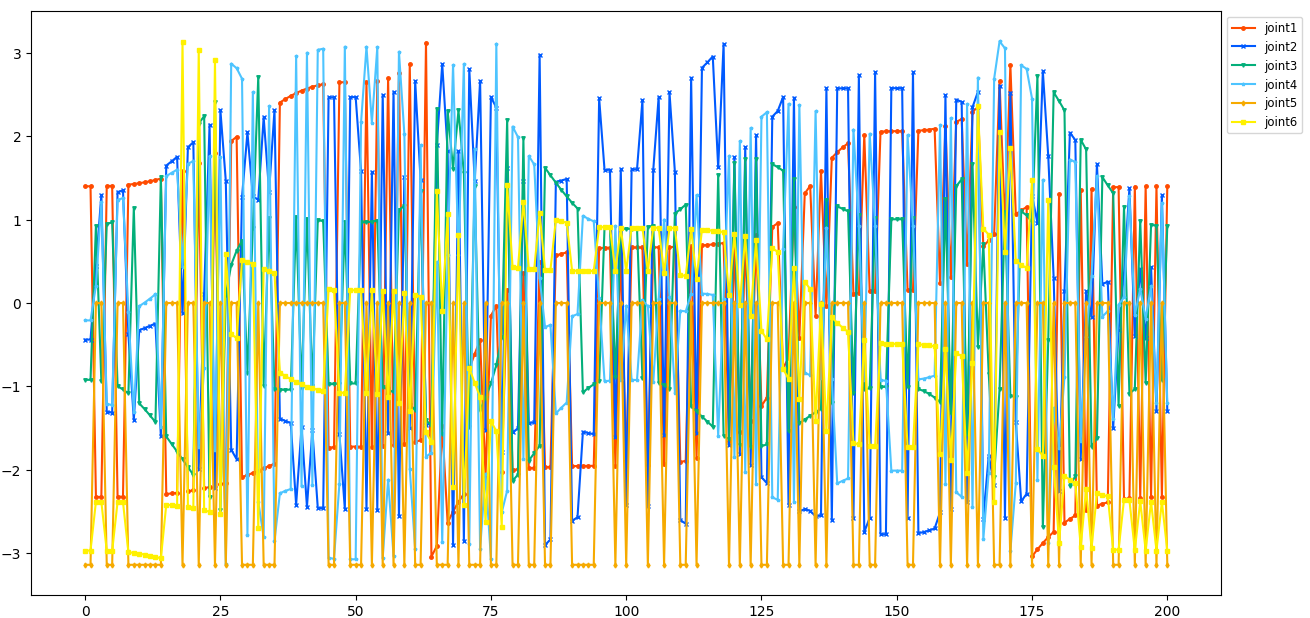}
  \caption{Displacements of joints during the operation of the end effector shown in Figure \ref{fig:Sm-50-all} (Test 6)}
  \label{fig:Sm-50-joint}
  \includegraphics[scale=0.2]{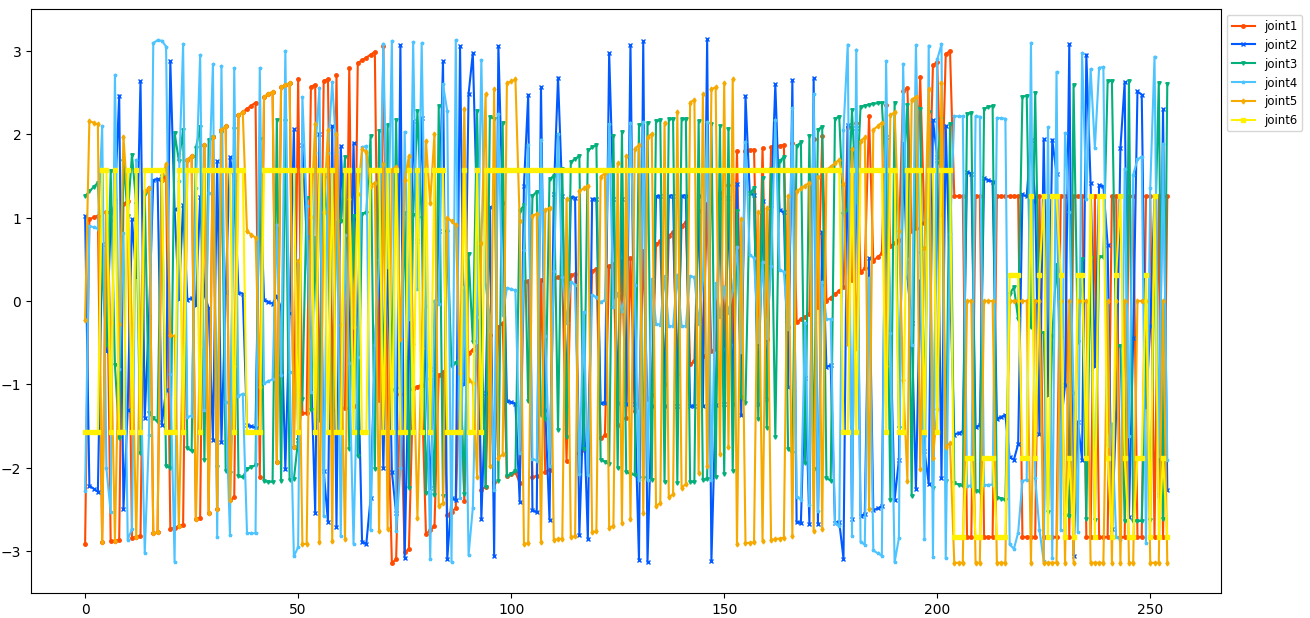}
  \caption{Displacements of joints during the operation of the end effector shown in Figure \ref{fig:Eq-50-ori} (Test 7)}
  \label{fig:Eq-50-joint}
\end{figure}
\cref{fig:Sm-50-joint,fig:Eq-50-joint} illustrate how each joint of myCobot is displaced at each via-point during the motion of the end-effector shown in \cref{fig:Sm-50-all,fig:Eq-50-ori}, respectively.
In the figures, the horizontal axis represents $t$, while the vertical axis represents the joint angle.
As these figures show, if one inverse kinematics solution is randomly selected at each via-point, the joint displacements between consecutive via-points may become large, potentially resulting in erratic overall motion of the myCobot. Therefore, it is necessary to select a sequence of inverse kinematics solutions such that the overall joint movements are minimized as much as possible.

In path optimization, the goal is to determine a sequence of joint configurations
that minimizes the overall motion of the myCobot in terms of joint displacement.

\subsection{Path Optimization Using Dijkstra's Algorithm}
\label{sec:path-optimization-dijkstra}

To solve the path optimization problem, we model the problem as a directed graph with weights on the edges, and apply Dijkstra's algorithm to find the optimal path.

Let $G=(V,E)$ be a directed graph, where $V$ is the set of vertices and $E$ is the set
of edges.
The vertices $V$ represent the inverse kinematics solutions at each via-point, and
the edges $E$ represent the transitions between these solutions.

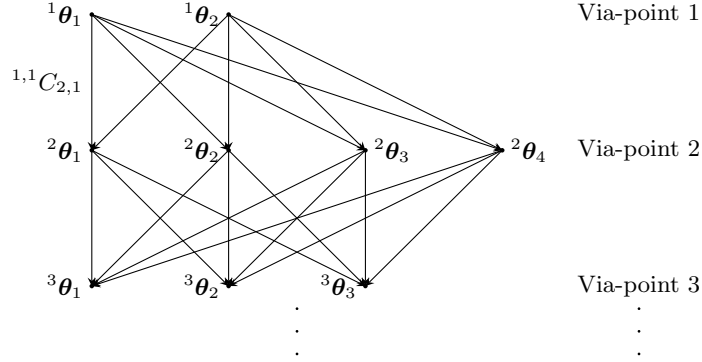
\begin{figure}[t]
  \centering
  \begin{tikzpicture}[scale=0.6]
    \fill (0,0) circle(0.05);
    \fill (3,0) circle(0.05);
    \fill (6,0) circle(0.05);
    \fill (0,3) circle(0.05);
    \fill (3,3) circle(0.05);
    \fill (6,3) circle(0.05);
    \fill (9,3) circle(0.05);
    \fill (0,6) circle(0.05);
    \fill (3,6) circle(0.05);

    \fill (4.5,-0.5) circle(0.03);
    \fill (4.5,-1) circle(0.03);
    \fill (4.5,-1.5) circle(0.03);
    \fill (12,-0.5) circle(0.03);
    \fill (12,-1) circle(0.03);
    \fill (12,-1.5) circle(0.03);

    \draw (0,0)node[left]{$^3\bm{\theta}_1$};
    \draw (3,0)node[left]{$^3\bm{\theta}_2$};
    \draw (6,0)node[left]{$^3\bm{\theta}_3$};
    \draw (0,3)node[left]{$^2\bm{\theta}_1$};
    \draw (3,3)node[left]{$^2\bm{\theta}_2$};
    \draw (6,3)node[right]{$^2\bm{\theta}_3$};
    \draw (9,3)node[right]{$^2\bm{\theta}_4$};
    \draw (0,6)node[left]{$^1\bm{\theta}_1$};
    \draw (3,6)node[left]{$^1\bm{\theta}_2$};

    \draw[arrows = {stealth[scale=3] -}] (0,0)--(0,3);
    \draw[arrows = {stealth[scale=3] -}] (0,0)--(3,3);
    \draw[arrows = {stealth[scale=3] -}] (0,0)--(6,3);
    \draw[arrows = {stealth[scale=3] -}] (0,0)--(9,3);
    \draw[arrows = {stealth[scale=3] -}] (3,0)--(0,3);
    \draw[arrows = {stealth[scale=3] -}] (3,0)--(3,3);
    \draw[arrows = {stealth[scale=3] -}] (3,0)--(6,3);
    \draw[arrows = {stealth[scale=3] -}] (3,0)--(9,3);
    \draw[arrows = {stealth[scale=3] -}] (6,0)--(0,3);
    \draw[arrows = {stealth[scale=3] -}] (6,0)--(3,3);
    \draw[arrows = {stealth[scale=3] -}] (6,0)--(6,3);
    \draw[arrows = {stealth[scale=3] -}] (6,0)--(9,3);
    \draw[arrows = {stealth[scale=3] -}] (0,3)--(0,6);
    \draw[arrows = {stealth[scale=3] -}] (0,3)--(3,6);
    \draw[arrows = {stealth[scale=3] -}] (3,3)--(0,6);
    \draw[arrows = {stealth[scale=3] -}] (3,3)--(3,6);
    \draw[arrows = {stealth[scale=3] -}] (6,3)--(0,6);
    \draw[arrows = {stealth[scale=3] -}] (6,3)--(3,6);
    \draw[arrows = {stealth[scale=3] -}] (9,3)--(0,6);
    \draw[arrows = {stealth[scale=3] -}] (9,3)--(3,6);

    \draw (12,0)node{Via-point 3};
    \draw (12,3)node{Via-point 2};
    \draw (12,6)node{Via-point 1};
    \draw (0,4.5)node[left]{${}^{1,1}C_{2,1}$};
  \end{tikzpicture}
  \caption{A directed graph representing displacements of joint configurations}
  \label{fig:Dijkstra}
\end{figure}

\cref{fig:Dijkstra} illustrates an example of the displacements of joint configurations in the directed graph.
Let
${}^i\bm{\theta}_{k}=({}^i\theta_{k,1},{}^i\theta_{k,2},{}^i\theta_{k,3},{}^i\theta_{k,4},{}^i\theta_{k,5},{}^i\theta_{k,6})$
denote the $k$-th inverse kinematics solution at the $i$-th via-point, where
$i=1,2,\dots,n$ and $k=1,2,\dots,m_i$.
The edges in the graph represent the displacements of the joint configurations
between two inverse kinematic solutions in the consecutive via-points.
For example, the edge from ${}^2\bm{\theta}_3$ to ${}^3\bm{\theta}_1$ represents
the displacement of joint configurations from the third inverse kinematics solution
at the second via-point to the first inverse kinematics solution at the third via-point.
Note that each edge is directed since, in the trajectory planning problem,
the displacement proceeds sequentially from via-point $i$ to via-point $i+1$.
Let ${}^{i,s}C_{i+1,t}$ denote the cost of the edge from ${}^i\bm{\theta}_s$
to ${}^{i+1}\bm{\theta}_t$, as shown in \cref{fig:Dijkstra}.

By using Dijkstra's algorithm, we can find the optimal path from the first via-point
to the last via-point as follows. Let the number of via-points be $n$, and the number
of inverse kinematics solutions at the $i$th via-point be $m_i$.
For $j=1,\dots,m_1$, we search for the shortest path from the solution of the inverse
kinematic problem ${}^j\bm{\theta}_1$ at the first via-point to the $l_j$th inverse
kinematics solution at the last via-point, denoted as ${}^{l_j}\bm{\theta}_n$, then
the shortest path is found by backtracking from the last via-point through the
previous via-points.
Then, by comparing the total costs of all paths, we find the optimal path from the first via-point to the last via-point.

\subsection{Definition of the Cost Functions}
\label{sec:cost-function}

In this section, we define the cost function used for the edges in Dijkstra's algorithm.
In our previous work \cite{shi-oka-ter-mik2024}, the sum of joint displacements was used
as the cost function. In this paper, in addition to that, we also consider the balance
of displacements across all joints and employ the following cost function.

We denote the cost function for the edge from ${}^i\bm{\theta}_s$ to
 ${}^{i+1}\bm{\theta}_t$ as
$f({}^i\bm{\theta}_s,{}^{i+1}\bm{\theta}_t)={}^{i,s}C_{i+1,t}$.

\subsubsection{The Sum of Joint Displacements}
The cost function $f_1$ is defined as
  \begin{equation}
    \label{eq:cost-sum}
    f_1({}^i \bm{\theta}_{s},{}^{i+1} \bm{\theta}_{t}) = \sum_{j=1}^6 |{}^{i}\theta_{s,j} -{}^{i+1} \theta_{t,j}|,
  \end{equation}
  which is the sum of the absolute values of the differences in joint angles between two consecutive via-points.
When this cost function is applied to Dijkstra’s algorithm, the resulting path
minimizes redundancy in terms of joint displacements, thereby reducing the overall
load on each joint. However, in some cases, a single joint may undergo
a significantly larger displacement, potentially leading to a concentration
of mechanical load.

\subsubsection{The Maximum Joint Displacement}
 The cost function $f_2$ is defined as
  \begin{equation}
    \label{eq:cost-max}
    f_2({}^i \bm{\theta}_{s},{}^{i+1} \bm{\theta}_{t}) = \max_{j \in \{1,\dots,6\}} |{}^{i}\theta_{s,j} -{}^{i+1}\theta_{t,j}|,
  \end{equation}
which is the maximum absolute value of the differences in joint angles between
two consecutive via-points.
When this cost function is applied to Dijkstra’s algorithm, the resulting path
minimizes the maximum joint displacement, thereby preventing excessive load on any
single joint.
However, since only the cost between adjacent via-points is considered, the total cost over the entire trajectory may exceed that obtained using the cost function $f_1$.

\subsubsection{The Standard Deviation of Joint Displacements}
 The cost function $f_3$ is defined as
\begin{equation}
    \label{eq:cost-standard-deviation}
    f_3({}^i \bm{\theta}_{s},{}^{i+1} \bm{\theta}_{t}) = \sqrt{\frac{1}{6}\sum_{i=1}^6 (\mathcal{J}_j - \overline{\mathcal{J}})^2},
\end{equation}
where $\mathcal{J}_j=|{}^{i}\theta_{s,j} -{}^{i+1}\theta_{t,j}|$ ($j=1,\dots,6$)
and $\overline{\mathcal{J}}=\frac{1}{6}\sum_{j=1}^6\mathcal{J}_j$,
which is the standard deviation of the joint displacements between two consecutive
via-points.
As a result, the generated path yields the most balanced joint displacements, which
is expected to equalize the mechanical load across all joints.
However, this cost function may also lead to cost equalization even in segments with
relatively large joint displacements, which could result in the selection of a path
that imposes a greater load on specific joints compared to those generated by other
cost functions.

\subsubsection{A Weighted Sum of the Above Three Cost Functions}
The cost function $f_4$ is defined as
  \begin{equation}
    \label{eq:cost-weighted-sum}
    f_4({}^i \bm{\theta}_{s},{}^{i+1} \bm{\theta}_{t}) = w_1 f_1({}^i \bm{\theta}_{s},{}^{i+1} \bm{\theta}_{t}) + w_2 f_2({}^i \bm{\theta}_{s},{}^{i+1} \bm{\theta}_{t}) + w_3 f_3({}^i \bm{\theta}_{s},{}^{i+1} \bm{\theta}_{t}),
  \end{equation}
where $w_1$, $w_2$, and $w_3$ are positive weights that satisfy $w_1+w_2+w_3=1$.
This cost function is a weighted sum of the three cost functions defined above,
allowing for a balance between minimizing the total joint displacement,
preventing excessive load on any single joint, and equalizing the mechanical
load across all joints.
The weights $w_1$, $w_2$, and $w_3$ can be adjusted according to the specific
requirements of the task at hand, such as prioritizing joint load balancing
or minimizing total joint displacement.
  In this paper, we set $w_1=0.4$, $w_2=0.2$, and $w_3=0.4$.

\subsubsection{The Manipulability Measure}
The manipulability measure \cite{yos1985} is a kinematic index used to quantitatively
evaluate the operational capability of a manipulator in controlling the position and
orientation of its end-effector.
Let $\bm{\theta}={}^t(\theta_1,\theta_2,\dots,\theta_n)$
be the joint angles of the manipulator
and let $\bm{r}={}^t(r_1,r_2,\dots,r_m)$ ($m\le n$) be the ``state'' of the 
end-effector, which consists of its position or orientation.
If the geometric relationship between $\theta$ and $r$ is expressed as
$r=f(\bm{\theta})$, then the manipulability $\omega$ is defined as
\begin{equation}
  \label{eq:manipulability}
  \omega = \sqrt{\det J(\bm{\theta})^t(J(\bm{\theta}))},
\end{equation}
where $J(\bm{\theta})$ is the Jacobian matrix of $f$ with respect to $\bm{\theta}$.
In this paper, we use $\bm{r}={}^t(p_1,p_2,p_3)=\bm{g}(\theta_1,\dots,\theta_6)$ 
where $p_i$ is the $i$th equation of \eqref{eq:CGS}, and the cost function $f_5$ 
is defined as
\begin{equation}
  \label{eq:cost-manipulability}
  f_5({}^i \bm{\theta}_{s},{}^{i+1} \bm{\theta}_{t}) =
  \sqrt{\det J({}^{i+1} \bm{\theta}_{t})^t(J({}^{i+1} \bm{\theta}_{t}))}.
\end{equation}
Note that, in \eqref{eq:cost-manipulability}, the cost function is computed 
at the inverse kinematics solution ${}^{i+1} \bm{\theta}_{t}$.

\subsubsection{The Weighted Sum of $f_4$ and $f_5$}
\label{sec:cost-weighted-sum-manipulability}

The cost function $f_4$ is used to optimize the displacements between each via-point. 
In contrast, the cost function $f_5$ is designed to optimize the orientation of the
robotic manipulator at each via-point along the path.
Therefore, by taking the weighted average of these cost functions, it is expected 
that a path can be generated in which the robotic manipulator maintains a favorable
orientation while minimizing and balancing joint displacements as much as possible.
The cost function $f_6$ is defined as
\begin{equation}
  \label{eq:cost-weighted-sum-manipulability}
  f_6({}^i \bm{\theta}_{s},{}^{i+1} \bm{\theta}_{t}) = w_4 f_4({}^i \bm{\theta}_{s},{}^{i+1} \bm{\theta}_{t}) + w_5 f_5({}^i \bm{\theta}_{s},{}^{i+1} \bm{\theta}_{t}),
\end{equation}
where $w_4$ and $w_5$ are positive weights that satisfy $w_4+w_5=1$.
In this paper, we set $w_4=1/(1+10^{-6})$ and $w_5=10^{-6}/(1+10^{-6})$ since 
the value of $f_4$ is approximately $O(10^1)$ while the value of $f_5$ is 
approximately $O(10^{7})$.

\subsection{Experiments}
\label{sec:experiments-path-optimization}

In this section, we conduct experiments to evaluate the effectiveness 
of the proposed path optimization method.

We have implemented the proposed method in Python.
The computing environment is the same as that used in \cref{sec:trajectory-planning}.
We use the same path as in the following tests in \cref{sec:trajectory-experiments}:
Test 1 with $T=25$ and $50$,
Test 2 with $T=25$ and $50$,
Test 3 with $T=25$ and $50$,
Test 4 with $T=25$ and $50$,
Test 6 with $T=25$ and $50$, and
Test 7 with $T=50$.

\subsubsection{Experimental Results on the Trajectory of Uniform Motion}
\label{sec:experiments-path-optimization-uniform}

In this section, we determined the trajectories of Tests 1, 2, 3, 4, 6, and 7 using the
method for uniform motion described in \cref{sec:trajectory-uniformly}, and then 
measured the computation time required to apply Dijkstra's algorithm to each 
trajectory for each cost function.

\cref{tab:Dijkstra-uniform-motion} shows the computing time of the experiments.
The first column indicates the number of via-points, and the second column indicates the test number.
The columns $f_1$ through $f_6$ represent the average computing time (in seconds) 
for each cost function.
The computing time for each cost function is expressed in the format 
``$a$e${b}$'', representing
$a \times 10^b$, and corresponds to the average of ten measurements taken 
for each trajectory.
The computing times are presented with three significant digits.
\begin{table}[t]
  \footnotesize
  \centering
  \begin{tabular}{c|c|cccccc}
  \hline
  T & Test & $f_1$ [s] & $f_2$ [s] & $f_3$ [s] & $f_4$ [s] & $f_5$ [s] & $f_6$ [s] \\
  \hline
  \multirow{5}{*}{25} & 1 & $8.42$e${-4}$ & $8.44$e${-4}$ & $1.61$e${-2}$ & $1.86$e${-2}$ &
   $7.71$e$1$ & $7.72$e$1$ \\ 
   &  2 & $7.54$e${-4}$ & $7.50$e${-4}$ & $1.20$e${-2}$ & $1.43$e${-2}$ & $6.82$e$1$ &
   $6.83$e$1$ \\ 
   &  3 & $4.69$e${-4}$ & $4.52$e${-4}$ & $6.96$e${-3}$ & $8.26$e${-3}$ & $3.99$e$1$ &
   $3.99$e$1$ \\ 
   &  4 & $7.79$e${-4}$ & $7.75$e${-4}$ & $1.23$e${-2}$ & $1.46$e${-2}$ & $7.01$e$1$ &
   $7.02$e$1$ \\ 
   &  6 & $2.66$e${-3}$ & $2.68$e${-3}$ & $4.36$e${-2}$ & $5.22$e${-2}$ & $2.54$e$2$ &
   $2.54$e$2$ \\ \hline
   \multirow{6}{*}{50} & 1 & $1.69$e${-3}$ & $1.70$e${-3}$ & $2.79$e${-2}$ & $3.35$e${-2}$ &
   $1.59$e$2$ & $1.60$e$2$ \\ 
   &  2 & $1.48$e${-3}$ & $1.48$e${-3}$ & $2.38$e${-2}$ & $2.84$e${-2}$ & $1.38$e$2$ &
   $1.38$e$2$ \\ 
   &  3 & $8.82$e${-4}$ & $8.72$e${-4}$ & $1.33$e${-2}$ & $1.59$e${-2}$ & $7.85$e$1$ &
   $7.86$e$1$ \\ 
   &  4 & $1.60$e${-3}$ & $1.61$e${-3}$ & $2.60$e${-2}$ & $3.12$e${-2}$ & $1.50$e$2$ &
   $1.50$e$2$ \\ 
   &  6 & $5.36$e${-3}$ & $5.44$e${-3}$ & $8.91$e${-2}$ & $1.08$e${-1}$ & $5.25$e$2$ &
   $5.25$e$2$ \\ 
   &  7 & $1.98$e${-2}$ & $1.99$e${-2}$ & $3.55$e${-1}$ & $4.29$e${-1}$ & $1.67$e$3$ &
   $1.67$e$3$ \\ \hline
  \end{tabular}
  \caption{The computing time for Dijkstra's algorithm for each cost function in the trajectory planning problem for uniform motion (see \Cref{sec:experiments-path-optimization})}
  \label{tab:Dijkstra-uniform-motion}
\end{table}

\begin{figure}[t]
  \begin{tabular}{cc}
    \begin{minipage}[t]{0.5\hsize}
      \centering
      \includegraphics[keepaspectratio, width = \linewidth]{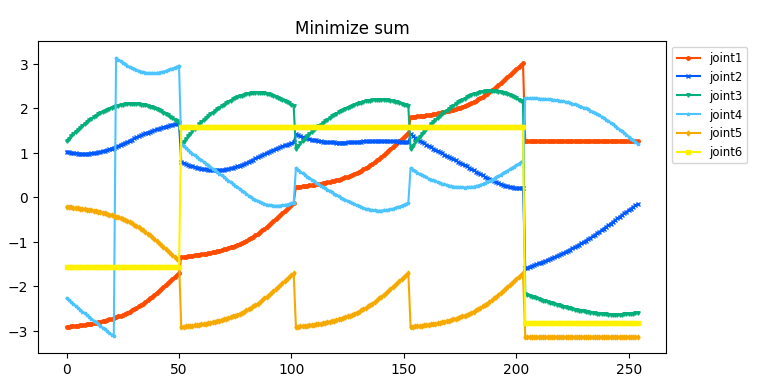}
      \subcaption{With cost function $f_1$}
      \label{fig:eq50ori-D1}
    \end{minipage} &
    \begin{minipage}[t]{0.5\hsize}
      \centering
      \includegraphics[keepaspectratio, width = \linewidth]{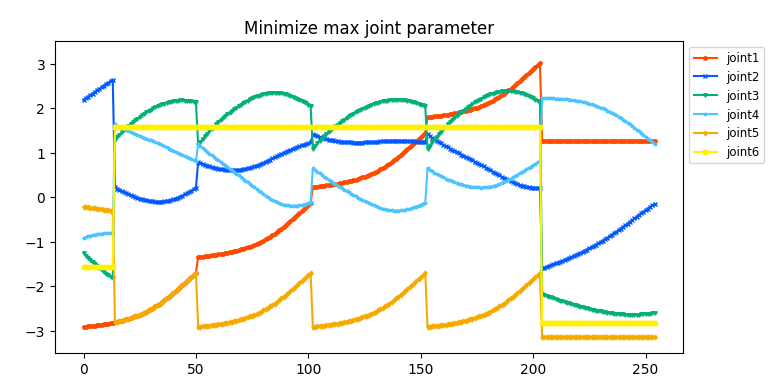}
      \subcaption{With cost function $f_2$}
      \label{fig:eq50ori-D2}
    \end{minipage} \\

    \begin{minipage}[t]{0.5\hsize}
      \centering
      \includegraphics[keepaspectratio, width = \linewidth]{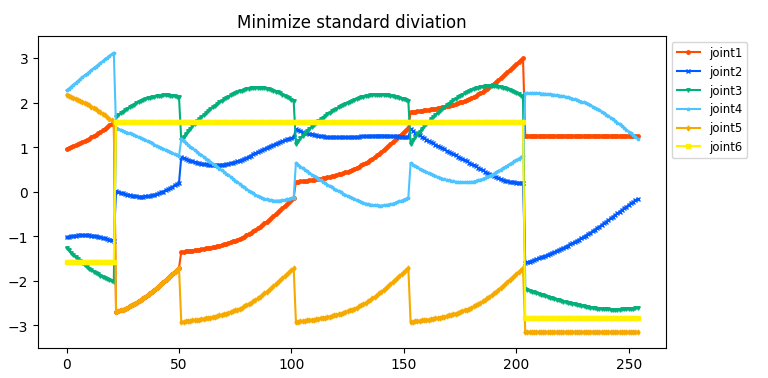}
      \subcaption{With cost function $f_3$}
      \label{fig:eq50ori-D3}
    \end{minipage} &
    \begin{minipage}[t]{0.5\hsize}
      \centering
      \includegraphics[keepaspectratio, width = \linewidth]{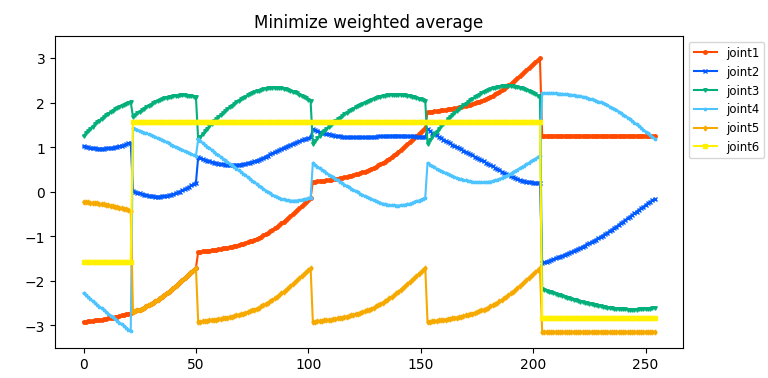}
      \subcaption{With cost function $f_4$}
      \label{fig:eq50ori-D4}
    \end{minipage}
  \end{tabular}
\caption{The results of Dijkstra's algorithm using the cost functions $f_1$ through $f_4$ (for the case of \cref{fig:Eq-50-joint}) (Test 7)}
\label{fig:eq50ori-D1234}
  \begin{tabular}{cc}
    \begin{minipage}[t]{0.5\hsize}
      \centering
      \includegraphics[keepaspectratio, width = \linewidth]{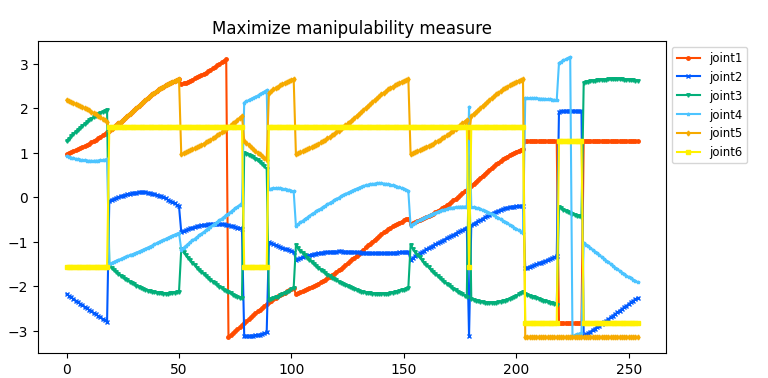}
      \subcaption{With cost function $f_5$}
      \label{fig:eq50ori-D5}
    \end{minipage} &
    \begin{minipage}[t]{0.5\hsize}
      \centering
      \includegraphics[keepaspectratio, width = \linewidth]{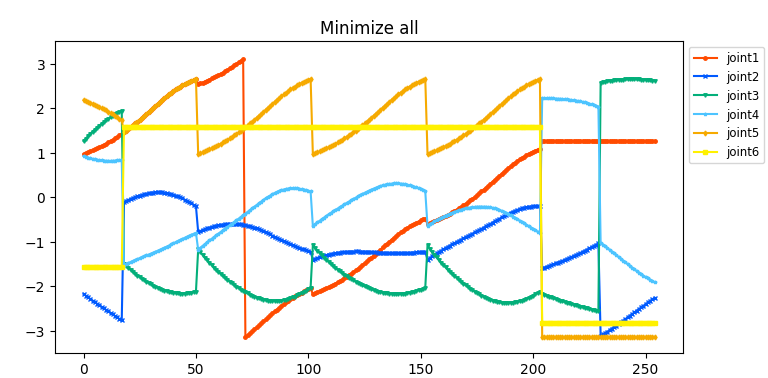}
      \subcaption{With cost function $f_6$}
      \label{fig:eq50ori-D6}
    \end{minipage}
  \end{tabular}
  \caption{The results of Dijkstra's algorithm using the cost functions $f_5$ and $f_6$ (for the case of \cref{fig:Eq-50-joint}) (Test 7)}
  \label{fig:eq50ori-D56}
\end{figure}

\cref{fig:eq50ori-D1234,fig:eq50ori-D56} show the results of Dijkstra's algorithm using
the cost functions $f_1$ through $f_6$ for Test 7, with the joint configuration 
resulting from a randomly selected solution to the inverse kinematics problem shown 
in \cref{fig:Eq-50-joint}.
\cref{fig:eq50ori-D1234}, \cref{fig:eq50ori-D1} through \cref{fig:eq50ori-D4}
show the results of Dijkstra's algorithm using the cost functions $f_1$ through $f_4$, 
respectively, \cref{fig:eq50ori-D5} and \cref{fig:eq50ori-D6} show 
the results using the cost functions $f_5$ and $f_6$, respectively.

\subsubsection{Experimental Results on the Trajectory of Smooth Motion}
\label{sec:experiments-path-optimization-smooth}

In this section, we determined the trajectories of Tests 1, 2, 3, 4, 6, and 7 using the
method for smooth motion described in \cref{sec:trajectory-smooth}, and then measured 
the computation time required to apply Dijkstra's algorithm to each trajectory for each
cost function.

\cref{tab:Dijkstra-smooth-motion} shows the computing time of the experiments.
The contents of the table and the precision of the computing times are the same as in
\cref{tab:Dijkstra-uniform-motion}.
\begin{table}[t]
  \footnotesize
  \centering
  \begin{tabular}{c|c|cccccc}
  \hline
  T & Test & $f_1$ [s] & $f_2$ [s] & $f_3$ [s] & $f_4$ [s] & $f_5$ [s] & $f_6$ [s] \\
  \hline
  \multirow{5}{*}{25} & 1 & $7.62$e${-4}$ & $7.57$e${-4}$ & $1.21$e${-2}$ & $1.45$e${-2}$ & $6.89$e${1}$ & $6.89$e${1}$ \\ 
   &  2 & $6.70$e${-4}$ & $6.59$e${-4}$ & $1.04$e${-2}$ & $1.25$e${-2}$ & $6.04$e${1}$ & $6.04$e${1}$ \\ 
   &  3 & $5.04$e${-4}$ & $4.88$e${-4}$ & $7.57$e${-3}$ & $8.98$e${-3}$ & $4.40$e${1}$ & $4.40$e${1}$ \\ 
   &  4 & $7.50$e${-4}$ & $7.41$e${-4}$ & $1.18$e${-2}$ & $1.41$e${-2}$ & $6.75$e${1}$ & $6.76$e${1}$ \\ 
   &  6 & $2.54$e${-3}$ & $2.52$e${-3}$ & $4.08$e${-2}$ & $4.91$e${-2}$ & $2.40$e${2}$ & $2.40$e${2}$ \\ \hline
   \multirow{6}{*}{50} & 1 & $1.45$e${-3}$ & $1.45$e${-3}$ & $2.32$e${-2}$ & $2.78$e${-2}$ & $1.34$e${2}$ & $1.34$e${2}$ \\ 
   &  2 & $1.35$e${-3}$ & $1.35$e${-3}$ & $2.15$e${-2}$ & $2.58$e${-2}$ & $1.26$e$2$ &
   $1.26$e${2}$ \\ 
   &  3 & $9.52$e${-4}$ & $9.37$e${-4}$ & $1.46$e${-2}$ & $1.75$e${-2}$ & $8.68$e${1}$ & $8.69$e${1}$ \\ 
   &  4 & $1.40$e${-3}$ & $1.40$e${-3}$ & $2.24$e${-2}$ & $2.68$e${-2}$ & $1.30$e${2}$ & $1.30$e${2}$ \\ 
   &  6 & $4.89$e${-3}$ & $4.92$e${-3}$ & $8.02$e${-2}$ & $9.67$e${-2}$ & $4.75$e${2}$ & $4.75$e${2}$ \\ 
   &  7 & $1.66$e${-2}$ & $1.69$e${-2}$ & $3.01$e${-1}$ & $3.63$e${-1}$ & $1.43$e${3}$ & $1.43$e${3}$ \\ \hline
  \end{tabular}
  \caption{The computing time for Dijkstra's algorithm for each cost function in the trajectory planning problem for smooth motion (see \Cref{sec:experiments-path-optimization})}
  \label{tab:Dijkstra-smooth-motion}
\end{table}

\begin{figure}[t]
  \begin{tabular}{cc}
    \begin{minipage}[t]{0.5\hsize}
      \centering
      \includegraphics[keepaspectratio, width = \linewidth]{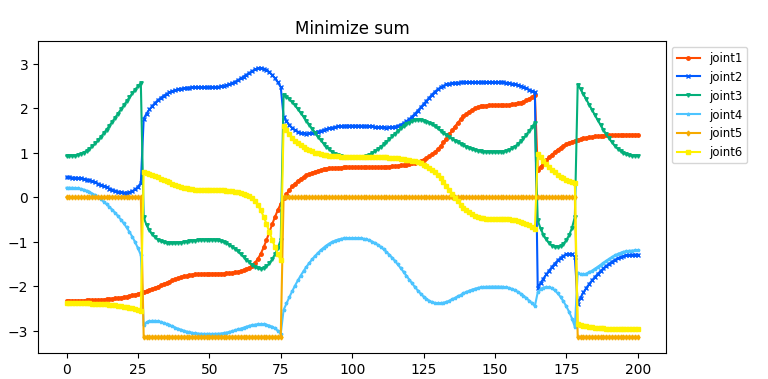}
      \subcaption{With cost function $f_1$}
      \label{fig:sm50all-D1}
    \end{minipage} &
    \begin{minipage}[t]{0.5\hsize}
      \centering
      \includegraphics[keepaspectratio, width = \linewidth]{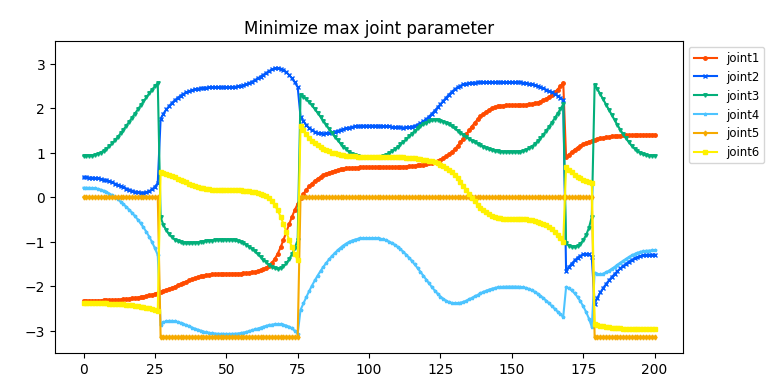}
      \subcaption{With cost function $f_2$}
      \label{fig:sm50all-D2}
    \end{minipage} \\

    \begin{minipage}[t]{0.5\hsize}
      \centering
      \includegraphics[keepaspectratio, width = \linewidth]{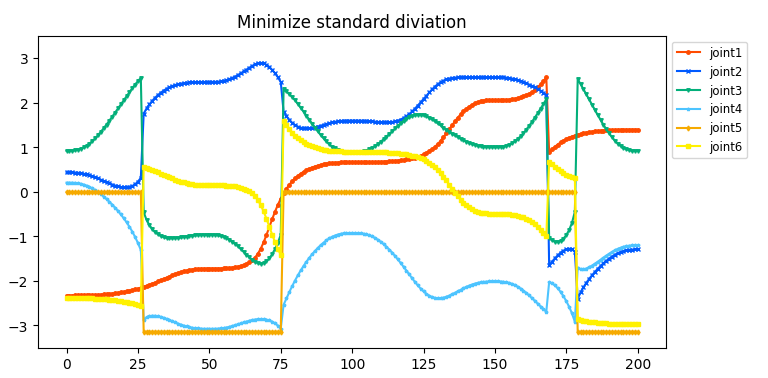}
      \subcaption{With cost function $f_3$}
      \label{fig:sm50all-D3}
    \end{minipage} &
    \begin{minipage}[t]{0.5\hsize}
      \centering
      \includegraphics[keepaspectratio, width = \linewidth]{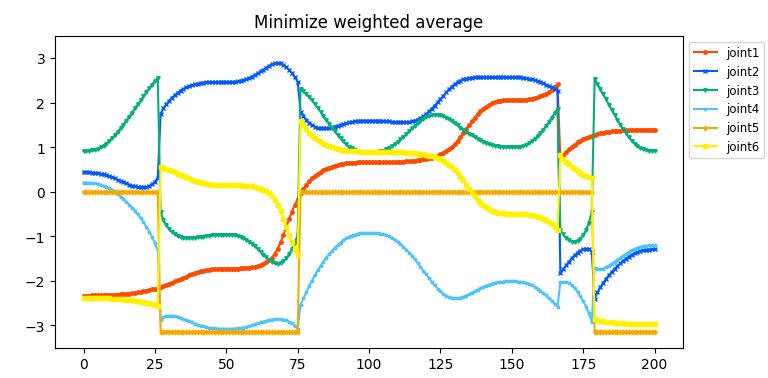}
      \subcaption{With cost function $f_4$}
      \label{fig:sm50all-D4}
    \end{minipage}
  \end{tabular}
\caption{The results of Dijkstra's algorithm using the cost functions through $f_1$ and $f_4$ (for the case of \cref{fig:Sm-50-joint}) (Test 6)}
\label{fig:sm50all-D1234}
  \begin{tabular}{cc}
    \begin{minipage}[t]{0.5\hsize}
      \centering
      \includegraphics[keepaspectratio, width = \linewidth]{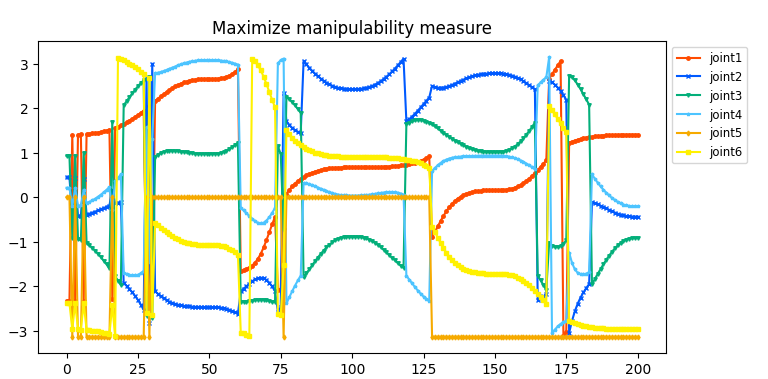}
      \subcaption{With cost function $f_5$}
      \label{fig:sm50all-D5}
    \end{minipage} &
    \begin{minipage}[t]{0.5\hsize}
      \centering
      \includegraphics[keepaspectratio, width = \linewidth]{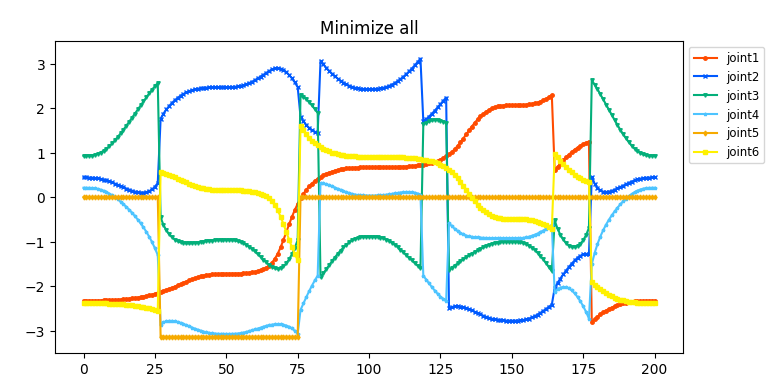}
      \subcaption{With cost function $f_6$}
      \label{fig:sm50all-D6}
    \end{minipage}
  \end{tabular}
  \caption{The results of Dijkstra's algorithm using the cost functions $f_5$ and $f_6$ (for the case of \cref{fig:Sm-50-joint}) (Test 6)}
  \label{fig:sm50all-D56}
\end{figure}

\cref{fig:sm50all-D1234,fig:sm50all-D56} show the results of Dijkstra's algorithm using 
the cost functions $f_1$ through $f_6$ for Test 6, with the joint configuration 
resulting from a randomly selected solution to the inverse kinematics problem 
is shown in \cref{fig:Sm-50-joint}.
\cref{fig:sm50all-D1234}, \cref{fig:sm50all-D1} through \cref{fig:sm50all-D4} show 
the results of Dijkstra's algorithm using the cost functions $f_1$ through $f_4$, respectively, while \cref{fig:sm50all-D5} and \cref{fig:sm50all-D6} show the results using the cost functions $f_5$ and $f_6$, respectively.

\subsection{Comparison and Analysis of Experimental Results}
\label{sec:comparison-analysis-experimental-results}
In this section, we compare the results of the experiments conducted 
in \cref{sec:experiments-path-optimization}.

\subsubsection{Comparison of Computing Times}

In this section, we compare the computing times across different trajectories 
and cost functions.

First, we discuss the computing time for each trajectory. The computing time of Dijkstra's algorithm appears to depend not on the number of via-points but rather on the number of solutions at each via-point.
\cref{tab:uniform-sol-number,tab:smooth-sol-number} show the number of solutions for each trajectory in \cref{sec:trajectory-uniformly} and \cref{sec:trajectory-smooth}, respectively.

In fact, by comparing \cref{tab:Dijkstra-uniform-motion,tab:uniform-sol-number}, 
as well as
\cref{tab:Dijkstra-smooth-motion,tab:smooth-sol-number}, we can observe that doubling 
the number of trajectory segments approximately doubles the total number of solutions.
Consequently, the computing time of Dijkstra's algorithm also increases by roughly a 
factor of two in proportion to the number of solutions.
In the case of T = 50 segments, although the difference in the number of via-points
between Test 6 and Test 7 is only 50, the total number of solutions in Test 7 is
approximately 2.5 times greater, resulting in a computing time for Dijkstra's algorithm
that is about three times longer.
Therefore, the computing time of Dijkstra's algorithm is considered to be 
approximately proportional to the number of such solutions.

\begin{table}[t]
  \begin{minipage}{0.45\textwidth}
  \centering
  \begin{tabular}{c|c|c}
    \hline
    T & Test & \# Solutions \\ \hline
    \multirow{5}{*}{25} & 1 & 126 \\ 
     &  2 & 118 \\ 
     &  3 & 90 \\ 
     &  4 & 122 \\ 
     &  6 & 444 \\ \hline
     \multirow{6}{*}{50} & 1 & 252 \\ 
     &  2 & 232 \\ 
     &  3 & 174 \\ 
     &  4 & 246 \\ 
     &  6 & 892 \\ 
     &  7 & 2018 \\ \hline
    \end{tabular}
    \caption{The total number of solutions for the proposed method in \cref{sec:trajectory-uniformly}}
    \label{tab:uniform-sol-number}
  \end{minipage}
  \hfill
  \begin{minipage}{0.45\textwidth}
    \centering
    \begin{tabular}{c|c|c}
      \hline
      T & Test & \# Solutions  \\ \hline\hline
      \multirow{5}{*}{25} & 1 & 120 \\ 
       &  2 & 112 \\ 
       &  3 & 96 \\ 
       &  4 & 120 \\ 
       &  6 & 436 \\ \hline
       \multirow{6}{*}{50} & 1 & 232 \\ 
       &  2 & 224 \\ 
       &  3 & 186 \\ 
       &  4 & 230 \\ 
       &  6 & 860 \\ 
       &  7 & 1840 \\ \hline
      \end{tabular}
    \caption{The total number of solutions for the proposed method in \cref{sec:trajectory-smooth}}
    \label{tab:smooth-sol-number}
  \end{minipage}
\end{table}

Next, we discuss the computing times for each cost function. The computing times 
follow the relationship:
$f_1, f_2 < f_3, f_4 \ll f_5, f_6$.
The cost functions $f_5$ and $f_6$, which are based on manipulability, tend to have
relatively high computing times. This is because they involve higher computational 
costs than the other cost functions, such as computing the Jacobian matrix and then 
its determinant.

\subsubsection{Comparison of the Results of Dijkstra's Algorithm}

In this section, we compare the results of Dijkstra's algorithm for each cost function.
\cref{fig:eq50ori-D1234,fig:sm50all-D1234} show that the cost functions
$f_1$, $f_2$, and $f_3$ generate similar paths; however, there are some sections 
where the paths differ.
In addition, the path $f_4$, which is derived from $f_1$, $f_2$, and $f_3$, is 
generally similar to those of $f_1$, $f_2$, and $f_3$, suggesting that the influence 
of each cost function varies depending on the location.
The fact that the path generated by cost function $f_4$ differs from all of $f_1$, 
$f_2$, and $f_3$, depending on the location, suggests that \( f_4 \) successfully produces a new path that incorporates elements of \( f_1 \), \( f_2 \), and \( f_3 \), as intended.
From \cref{fig:eq50ori-D56,fig:sm50all-D56}, it can be observed that the cost function
\( f_5 \), which is based on manipulability, contributes to generating more complex 
paths compared to \( f_1 \), \( f_2 \), \( f_3 \), and \( f_4 \).
This is because, unlike the other cost functions, manipulability does not focus on 
joint displacement. However, from a local perspective, the generated paths appear 
more natural compared to those obtained by randomly selecting a single solution. 
This suggests that when via-points are close to each other, joint configurations that 
are far from singularities and yield higher manipulability are also likely to be close 
to one another.
On the other hand, the path generated by the cost function \( f_6 \), which is derived
from \( f_4 \) and \( f_5 \), is smoother than that of \( f_5 \). This suggests that 
the path generated by \( f_6 \) achieves more natural joint displacements while 
also taking manipulability into account.

\section{Concluding Remarks}
\label{sec:conclusion}

In this paper, we proposed a solution to the trajectory planning problem in which the
end-effector of a 6-DOF robot manipulator, myCobot, moves along a straight line while
maintaining a fixed orientation for the end-effector, based on the analytical solutions
obtained from the inverse kinematics of the manipulator using computer algebra.
In this process, there may be several inverse kinematics solutions for a single 
via-point on the trajectory. To ensure smooth motion of the manipulator, it is necessary
to select one solution for each via-point and generate a unique path.
To select a unique path, we formulated the problem as a path optimization problem, 
applied Dijkstra's algorithm, and proposed various cost functions to optimize the path.
We then derived methods to obtain optimal paths that satisfy different conditions.

The following points are identified as future research challenges.
The first challenge is to demonstrate that the proposed method is efficient.
For this purpose, it is necessary to compare it with existing methods for trajectory
planning and path optimization that are based on computer algebra and numerical 
algorithms.

The second issue concerns trajectory planning with more complicated trajectories, 
such as curved paths.
In this paper, for the sake of simplicity, we address the trajectory planning problem 
in which the end-effector moves along a straight line. However, linear trajectories
present two main challenges. The first arises when obstacles are present along the
straight path.
In such cases, the end-effector must generate a detoured trajectory to avoid the 
obstacles.
The second case involves combining multiple linear trajectories, as in Test 6. In 
such situations, the direction of acceleration of the end-effector changes abruptly 
at the waypoints between trajectories, resulting in large forces being applied to 
the end-effector and increasing mechanical stress. Therefore, it is necessary to 
generate curved trajectories that ensure smooth transitions at these junctions.
As potential solutions to these issues, spline curves proposed by Shirato et al.
\cite{shi-oka-ter-mik2024}
and B\'ezier curves proposed by Hatakeyama et al. \cite{hat-ter-mik2024}
have been utilized for obstacle-avoiding trajectories. In addition, clothoid curves
\cite{che-cai-zhe2017}
have been suggested as a means of smoothly connecting linear segments.

The third issue concerns the formulation of an optimal cost function for the path
optimization problem.
In this paper, a weighted average was employed to construct a composite cost function 
by combining multiple cost functions within the framework of Dijkstra’s algorithm for 
the path optimization problem.
When using a weighted average, if one of the combined cost functions takes significantly
smaller values compared to the others, the influence of the remaining cost functions
becomes dominant, and the contribution of the cost function with smaller values is 
likely to be ignored.
However, in the context of Dijkstra’s algorithm, where smaller costs lead to more 
optimal paths, it can be argued that lower cost values should be given higher priority.
Therefore, when constructing a composite cost function from multiple individual cost 
functions, it may be effective to define it in such a way that smaller values are 
given higher priority. This approach warrants further investigation in future work.

\section*{Acknowledgments}
  The authors would like to thank the anonymous reviewers for their helpful comments.
  This research is partially supported by JKA and its promotion funds from KEIRIN RACE.

\bibliographystyle{splncs04}
\bibliography{casc-2025-mycobot-path-planning}

\end{document}